\documentclass[3p]{elsarticle}

\usepackage{hyperref}

\hypersetup{
    colorlinks=true,
		citecolor= blue,
    linkcolor=red,	
    urlcolor=cyan,
    bookmarks=true}

\usepackage{graphicx}
\usepackage{longtable}
\usepackage{newtxtext,newtxmath}
\usepackage{booktabs}
\usepackage{amsmath}
\usepackage[
  separate-uncertainty = true,
  multi-part-units = repeat
]{siunitx}

\usepackage{geometry} % just not to bother with the table width

\usepackage{float}
\usepackage{lscape}

\author[rvt]{Rumaisah Munir\corref{cor1}}
\author[rvt,focal]{Rizwan Ahmed Khan\corref{cor2}}
\cortext[cor1]{Corresponding author}
\cortext[cor2]{Principal corresponding author}
\address[rvt]{Faculty of IT, Barrett Hodgson University, Karachi, Pakistan.}
\address[focal]{LIRIS, Universit\'e Claude Bernard Lyon1, France.}

\begin{document}
\begin{frontmatter}

\title{An Extensive Review on Spectral Imaging in Biometric Systems: Challenges \& Advancements}

%\maketitle
\begin{abstract}
Spectral imaging has recently gained traction for face recognition in biometric systems. We investigate the merits of spectral imaging for face recognition and the current challenges that hamper the widespread deployment of spectral sensors for face recognition. The reliability of conventional face recognition systems operating in the visible range is compromised by illumination changes, pose variations and spoof attacks. Recent works have reaped the benefits of spectral imaging to counter these limitations in surveillance activities (defence, airport security checks, etc.). However, the implementation of this technology for biometrics, is still in its infancy due to multiple reasons. We present an overview of the existing work in the domain of spectral imaging for face recognition, different types of modalities and their assessment, availability of public databases for sake of reproducible research as well as evaluation of algorithms, and recent advancements in the field, such as, the use of deep learning-based methods for recognizing faces from spectral images. 

%\textit{Key words:} Spectral imaging, biometrics, face recognition, spoof attacks, Deep learning. 
\end{abstract}
\begin{keyword}
\texttt{Spectral imaging \sep biometrics \sep face recognition \sep spoof attacks \sep Deep learning}

\end{keyword}
\end{frontmatter}

\section {Introduction}

\begin{figure*}[!htb]
\centering
\includegraphics[scale=0.5]{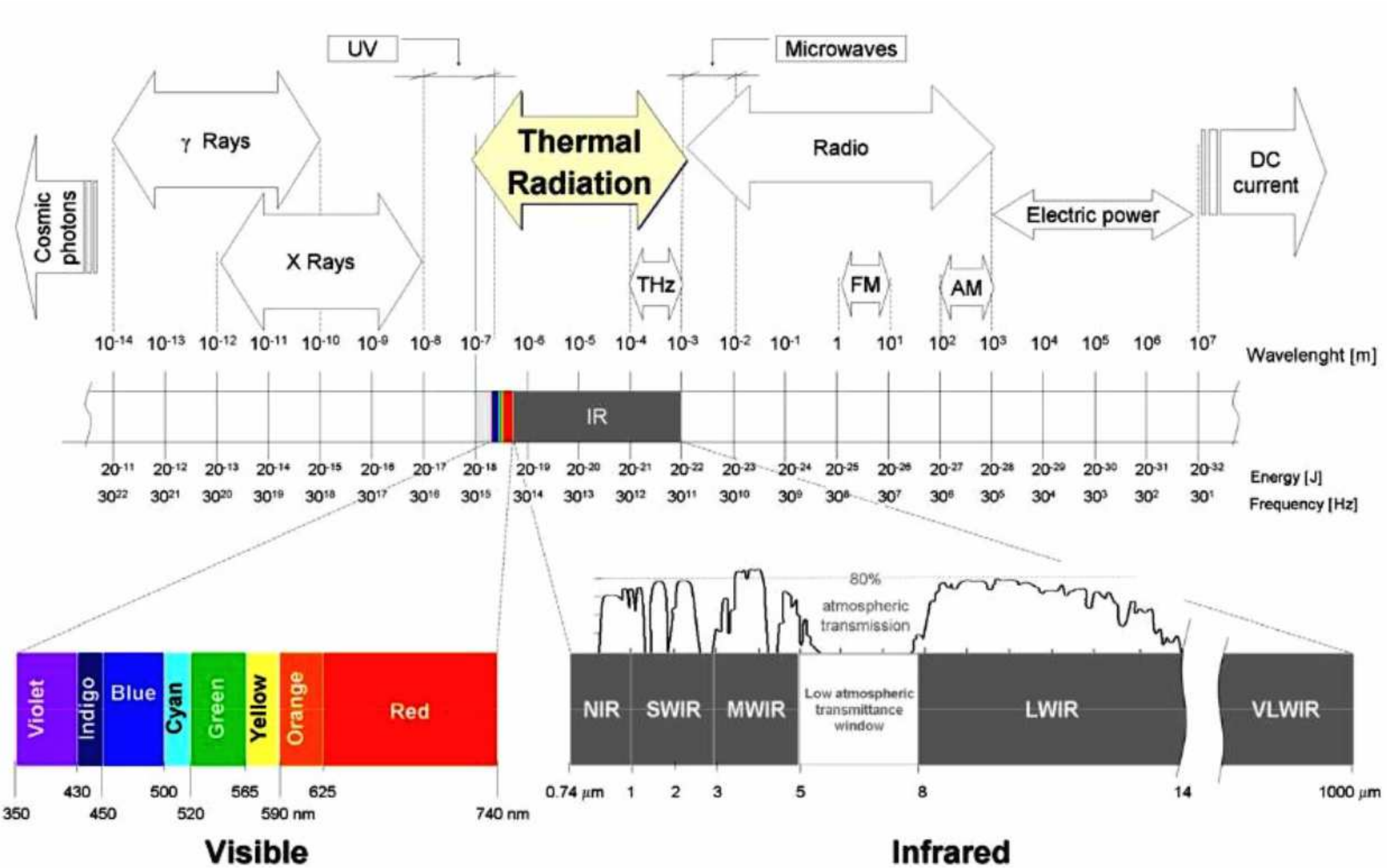}
\caption{Electromagnetic Spectrum: Bands and their wavelengths \cite{ibarra2005quantitative}} \label{pansy}
\end{figure*}

Spectral imaging refers to acquisition of images in various bands or sub-bands of the electromagnetic spectrum, for extraction of additional information of an object beyond that which is captured by human vision. Our survey is based on the use of spectral imaging for biometric systems, the challenges involved and recent advancements. A biometric system uses human physiological features for individual identification. Current biometric systems operate in the visible range of the electromagnetic spectrum. The electromagnetic spectrum consists of a range of wavelengths and corresponding frequencies. For visible spectrum in which the current biometric systems operate, the range is from 350nm to 740nm as depicted in Figure \ref{pansy}. 

Recently, spectral imaging gained traction for face recognition since there is information that can be exploited at specific wavelengths of the electromagnetic spectrum, or in other modality (such as InfraRed), which gets lost in conventional broadband imaging where information is averaged out (such as in RGB images, as seen in literature \cite{PhDthesis}). Each image obtained at a specific wavelength represents different information compared to another image captured at another wavelength. This discriminatory information can be exploited for higher recognition rates in different settings. Current biometric systems for face recognition operating in the visible range (broadband imaging) are compromised due to illumination conditions \cite{changfusion}, human pose variations, human facial expression variations and presentation attacks at the spectral sensor \cite{Ramachandra:2017:PAD:3058791.3038924}.

Spectral imaging has been around for over a decade. It has been used for cell microscopy \cite{zimmermann2003spectral}, breast imaging \cite{shikhaliev2011photon}, food engineering \cite{wu2008application}, cancer detection \cite{farkas2001applications}, eye imaging \cite{cabib2003spectral}, industrial applications \cite{hyvarinen1998direct} and remote sensing \cite{shaw2003spectral}.  

Application of spectral imaging for face recognition has shown promising results in the research domain \cite{4712365, changfusion}, catering for limitations such as facial expression variations, pose variations and presentation attacks \cite{Bouchech:2015:KSA:2821861.2821895, steiner2016reliable, raghavendra2017face, raghavendra2017vulnerability, vetrekar2017extended, 6613019}. Despite such promising results, smaller size and lower cost of the sensors used \cite{Allen16}, spectral imaging has not been deployed at large scale in biometric systems, yet.

\subsection{A generic Biometric System}
\begin{figure}
\centering
\includegraphics[scale=0.6]{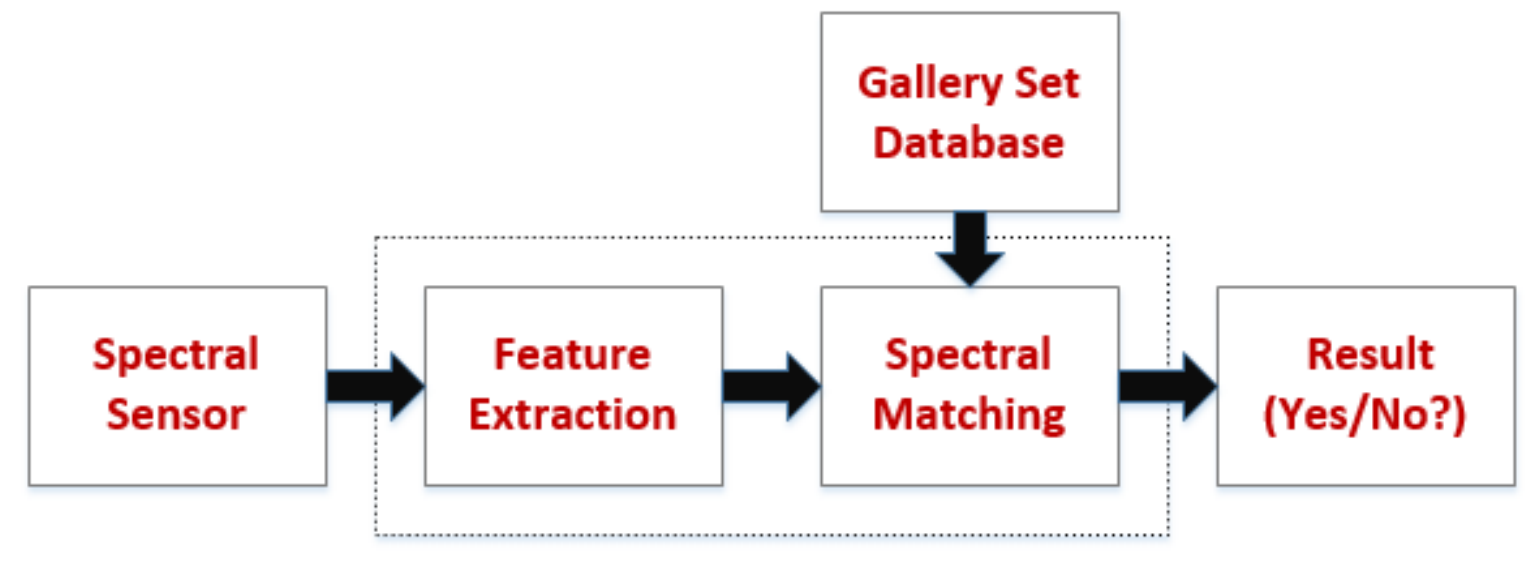}
\caption{A generic biometric system} \label{pink}
\end{figure}

A generic biometric system can be seen in Figure \ref{pink}. In a biometric system such as shown in Figure \ref{pink}, a unique physiological or behavioural feature, is used for personal identification of an individual. The system processes samples that record these physiological or behavioural features \cite{Li:2011:HFR:2073486}. Some biometric traits that have been used thus far, are iris \cite{boyce2006multispectral, ross2009exploring}, face \cite{6284307, 4563054}, dorsal hand \cite{zhang2016multiple}, fingerprint \cite{nixon2005multispectral, rowe2005multispectral}, palmprint \cite{zhang2010online, guo2012feature} etc. Soft biometric traits such as gender, height, age, etc. have also been combined to obtain better recognition results or act as preprocessing steps to make the recognition process computationally efficient \cite{jain2004soft, park2010face}.  

Two sets of images are required at least, to make a biometric system operational. One set consists of images of individuals that are known to the system. A process called feature extraction helps in reducing the information in these images to a set of features, which are then saved as templates, in the gallery set (the database of images or training images). Gallery set images are obtained in the enrolment phase, where the trait of an individual is enrolled in the system which is captured by a sensor. The other set of images consists of images of individuals that are presented to the biometric system for personal identification. These images are also converted to respective templates by feature extraction. This set is called the probe set (set of testing images). In the recognition stage the sensor captures the trait and after extraction of features, a comparison is made with the template stored during the enrolment phase \cite{1262027}. This comparison is made either by template matching or classification. The system accepts or rejects the result of the process. Classification methods are used to learn from the extracted features in the gallery set and with the help of those, classify the probe image.  

\begin{figure*}[htb]
\centering
\includegraphics[scale=1]{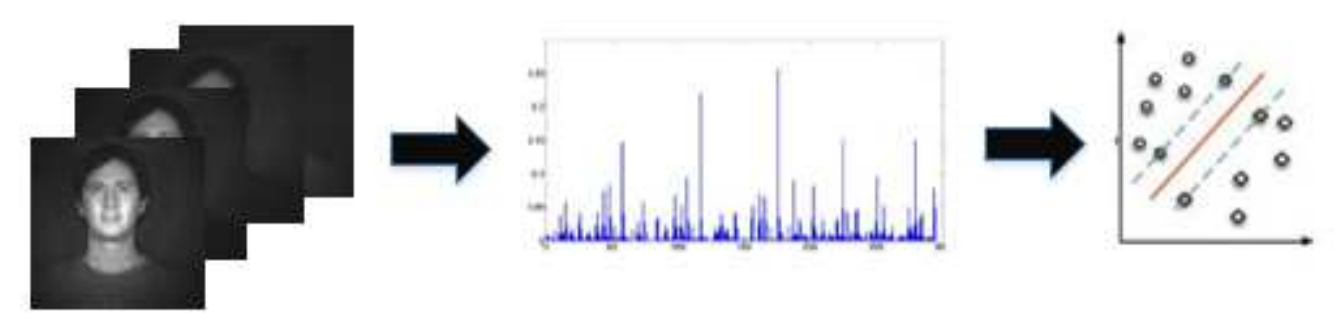}
\caption{A pipeline for face recognition algorithms \cite{khanriz}} \label{white}
\end{figure*}

Figure \ref{white} shows the steps involved in the analysis of a face image \cite{khanriz}. The first step requires the detection of a face in the image. Feature selection and extraction take place in the second step. Classification with a suitable classifier is the last step. As discussed earlier, a biometric system will either accept or reject the probe image or template. Sometimes the system is subject to fraud and falsely accepts an impostor disguised as an authorized user. The disguise can be of various forms such as beauty cosmetics, fake beard, a 3D mask or a printed photo \cite{Ramachandra:2017:PAD:3058791.3038924, 8244390, 7194916, 6909967}. It is formally called an \textit{artifact} with the help of which the attack is carried out. As a result, the biometric system is compromised. Therefore, one important consideration is to choose the right biometric trait, for which producing an \textit{artifact} may prove a challenge. The next section deals with the selection of a biometric trait from among many different physiological and behavioural traits of an individual. 

\subsection{Selection of a Biometric Characteristic}
From among the various traits that can be used for recognition, face is that common biometric characteristic which has a high universality, collectability, circumvention and acceptability \cite{1262027}. As per \cite{1262027}, the following requirements should be met by a biometric characteristic for it to qualify for use:

\begin{itemize}

\item Universality (U): Every individual must possess this trait.
\item Distinctiveness (D): Two individuals should be distinguishable with the help of this trait.
\item Permanence (Perm): This trait should not alter with time. 
\item Collectability (Co): It should be easily obtainable.  
\item Performance (Perf): The trait should help in achieving the desired accuracy.
\item Acceptability (A): The trait should be such that individuals willingly allow the use of this trait as an identifier.
\item Circumvention (Ci): This accounts for how easily an \textit{artifact} for the chosen trait can be produced .

\end{itemize} 

In Table \ref{bbb}, we have chosen the biometric identifiers that are most relevant to the domain of spectral imaging of individuals for the purpose of identification. We show which traits are favorable to qualify for use. We have chosen face as the most favorable biometric trait and based our survey on it due to it's high collectability and acceptability. It is also the most commonly used biometric trait. For spectral imaging of the other identifiers, we refer the reader to \cite{Zhang:2016:MBS:3126463}.

\begin{table*}[!h]
\centering
\begin{tabular}{ | l | l | l | l | l | l | l | l | }
\hline
	\textbf{Biometric Identifier} & \textbf{U} & \textbf{D} & \textbf{Perm} & \textbf{Co} & \textbf{Perf} & \textbf{A} & \textbf{Ci} \\ \hline
	\textbf{Face} & High & Low & Medium & High & Low & High & High \\ \hline
	\textbf{Facial thermogram} & High & High & Low & High & Medium & High & Low \\ \hline
	\textbf{Fingerprint} & Medium & High & High & Medium & High & Medium & Medium \\ \hline
	\textbf{Hand vein} & Medium & Medium & Medium & Medium & Medium & Medium & Low \\ \hline
	\textbf{Hand geometry} & Medium & Medium & Medium & High & Medium & Medium & Medium \\ \hline
	\textbf{Iris} & High & High & High & Medium & High & Low & Low \\ \hline
	\textbf{Palmprint} & Medium & High & High & Medium & High & Medium & Medium \\ \hline
\end{tabular}
\caption{In this table, we have reproduced the biometric identifiers relevant for spectral imaging as used by Jain \textit{et al.} \cite{1262027}. \newline The details of the \textit{initialisms} are as follows:
\newline U: Universality, D: Distinctiveness, Perm: Permanence, Co: Collectability, Perf: Performance, A: Acceptability, Ci: Circumvention} \label{bbb}
\end{table*}

Once the trait has been selected, a database can be collected for that trait for all individuals in a biometric system. The most common traits used are shown in Table \ref{bbb}. Face can be captured easily in most conditions. It is a unique trait which does not change with time once an individual reaches adulthood. Another such trait is the iris. An iris presents variations due to different colors in different human beings. Fingerprint and palmprint vary for all individuals due to variation in the pattern of fine lines. Although these traits have been used for biometric systems, only face and fingerprint based systems have been widely deployed around the world. Further to this, in case of spectral imaging most datasets in any other modality or wavelength, are developed largely for face rather than for fingerprint. 

We focus our survey on face as the biometric trait and study the challenges present in current systems for face recognition. In this survey, section \ref{headingface} deals with spectral imaging of face. In section \ref{promisingIR}, we focus our attention on other spectral bands that can be used to exploit information of the face beyond human vision. Section \ref{publicdata} contains the details of widely used publicly available databases. Section \ref{quant} deals with a quantitative analysis of different methods applied to the cited datasets in section \ref{publicdata}. Section \ref{spectralmatch} deals with literature work with regards to the matching process of images in the gallery set (database images/training images) with those in the probe set (testing images). Section \ref{apIR} highlights some of the important works in the literature of active and passive InfraRed. For further reading on active and passive InfraRed, refer to section \ref{promisingIR}. Section \ref{spoofsection} takes on tackling the challenges of current biometric systems for face recognition which suffer from spoof attacks, with the help of spectral imaging technology. Section \ref{neuralsection} highlights the use of deep learning for face recognition using spectral images. Conclusion and future direction are both presented in section \ref{futuresection}. 

\section {Spectral Imaging of Face} \label{headingface}

The spectral imaging of face is favorable in terms of the ease with which face can be captured with spectral sensors compared to other traits. Face is a universal trait. For the purpose of surveillance and security, face is that biometric trait which is most commonly captured, hence it has high collectability. 

As stated earlier, spectral imaging of face involves using different spectral bands of the electromagnetic spectrum to gather information in those particular bands. The term spectral imaging in our survey, includes both multispectral imaging and hyperspectral imaging. The former has the ability to image a scene in tens of spectral bands whereas the latter has the ability to image a scene in hundreds of spectral bands \cite{Allen16, activespectralimaging}. The results of a spectral sensor depend notably on the optical properties of the objects in the scene and the illumination geometry. These sensor results have to be converted to reflectance units. The result is a reflectance spectrum which can also be referred to as spectral signature of the objects in the image. It depends on the composition of the object and helps in classification \cite{activespectralimaging}. These spectral signatures are the fingerprints needed to differentiate objects in images. 

The reflectance spectrum of human skin has been studied by Angelopoulou \textit{et al.} \cite{Angelopoulou99thereflectance}. Experiments were conducted by Angelopoulou \textit{et al.} \cite{Angelopoulou99thereflectance} to establish how reflectance spectrum can help in distinguishing between skin and non-skin regions. According to Angelopoulou \textit{et al.}, since the human skin is capable of absorbing light, a non-skin region can be distinguished from a skin region. The experimental facility was designed to measure the reflectance spectrum of visible light reflected by an object's surface. 

Light reflected from the surface of an object depends on the following factors:

\begin{enumerate}
\item Direction of light
\item Light intensity
\item Spectral band

\end{enumerate}

The experimental facility used by Angelopoulou \textit{et al.} measured the \textit{Bidirectional Reflectance Distribution Function} (BRDF) of different samples. BRDF is a mathematical function that quantifies reflectance using five variables to measure light reflected from the surface of an object. These five variables are wavelength, and spherical coordinates for both, angle of incidence and angle of reflectance. 

Reflectance spectrum of skin was measured for 23 test subjects (18 males and 5 females, age range: 20-40 years old). There was a variety in the skin tones of the subjects chosen for the study. There were Asians, Caucasians, Africans and Indians. An important observation from the results of BRDF plots was that the darker skin reflected a small amount of incident light. These experiments were conducted for the skin of the hand and a mannequin to distinguish between skin regions and non-skin regions. The reflectance plots for both had considerable variations at different wavelengths. 

Uzair \textit{et al.} \cite{uzairbiometric} also conducted experiments to verify if the spectral reflectance profile of face really made a reliable biometric for individual identification, based on measurements. These measurements were obtained using a high precision spectrometer. A new database was acquired and made publicly available for reproducible results called UWA-FSRD. After testing their dataset against two other datasets (UWA-HSFD \& CMU HSFD), they concluded that identification rates of skin reflectance spectrum were not satisfactory. Details of the two datasets UWA-HSFD \& CMU HSFD can be found in section \ref{publicdata}.

Nevertheless, we investigate how spectral imaging, specially sub-bands in InfraRed region have been used to gain better recognition results with face as the characteristic trait. 

\section{Spectral Bands for Face Recognition} \label{promisingIR}

Current systems make use of images captured in the visible range (350-740 $\eta$m) of the electromagnetic spectrum \cite{10.1117/12.918899}. However, these systems suffer from degradation of performance due to varying illumination conditions. InfraRed is another modality that can be used by current systems. The advantage it has over visible is it's invariance to ambient lighting \cite{Sarfraz2015DeepPM}. Figure \ref{pansy}, shows bands in the electromagnetic spectrum and their corresponding wavelengths. We see that the range of the InfraRed (IR) radiation band in the electromagnetic spectrum begins from the red edge of the visible spectrum to 1 millimeter (mm). The InfraRed spectrum is composed of the active IR band and the passive (also referred to as thermal) IR  band \cite{6284307}. We observe the wavelengths of the these bands in Table \ref{xxx}.

The active IR band is reflection dominant and is further composed of Near-IR (NIR) and Short-Wave IR (SWIR). The NIR face imagery helps in tackling illumination direction changes and low-light conditions \cite{6710158}. SWIR has the advantage of being resistant to fog and smoke. Another advantage is that the SWIR imaging system is unobtrusive due to the fact that the reflected IR light is invisible to the eye \cite{6117586}. This is why SWIR is suited for surveillance activities. NIR and SWIR are closer to the visible spectrum.

The passive or thermal  IR  band is emission dominant and is further composed of the Mid-Wave InfraRed (MWIR) and the Long-Wave InfraRed (LWIR) bands. In the absence of a light source, thermal imagery is successful in capturing thermal signatures or thermograms of the skin tissue with the help of thermal sensors. The range of MWIR is 3-5 $\mu$m while that of LWIR is 8-14 $\mu$m. Since both MWIR and LWIR are thermal bands, their cameras sense variations in the facial temperature and as a result, produce facial thermograms. These thermograms are in the form of 2D images \cite{10.1117/12.918899}. Both MWIR and LWIR are thermal bands, sensing temperature variations but there is a difference between the two. MWIR has both emissive and reflective properties, whereas LWIR only has the former. MWIR is beneficial because no external illumination is needed to acquire MWIR images while a light source is needed for the active IR imagery. MWIR can detect vein patterns which helps provide a better overall biometric trait. LWIR has the advantage of being able to operate in the dark. According to Mendez \textit{et al.} \cite{10.1007/978-3-642-01793-3_34}, due to high emissivity of the human skin in the LWIR range, a thermal signature is obtained.

\begin{figure}
\centering
\includegraphics[scale=0.65]{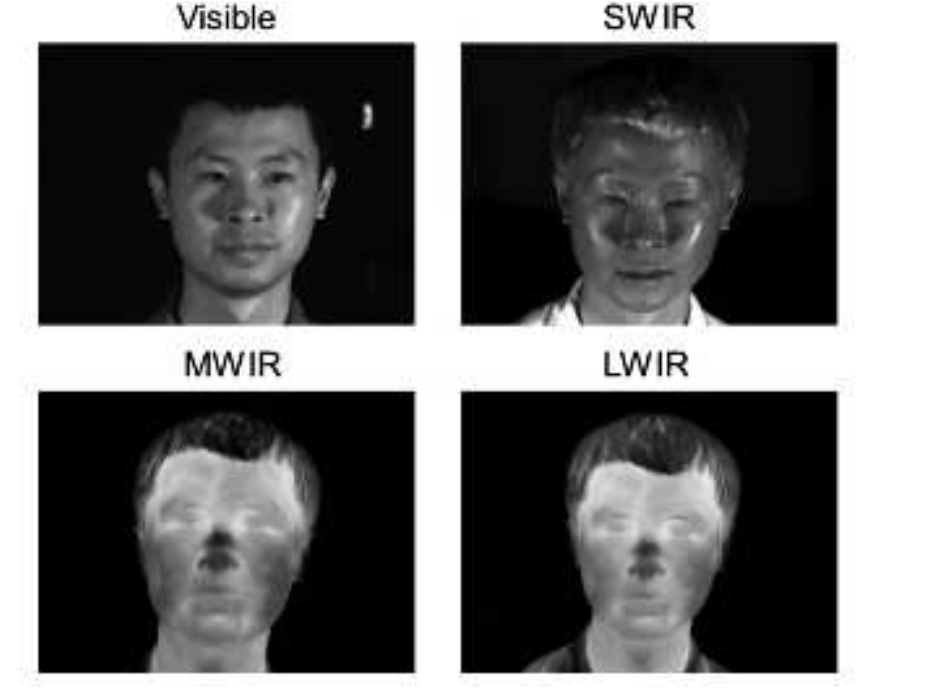}
\caption{Figure reproduced from the work of Hu \textit{et al.} \cite{Hu16}: Facial signatures of an individual acquired in different modalities simultaneously} \label{frog}
\end{figure}

Figure \ref{frog}, which has been reproduced from the work of Hu \textit{et al.} \cite{Hu16}, shows face images obtained in four different modalities i.e the visible range, SWIR, MWIR and LWIR. Apart from these bands, various sub-bands in the visible spectrum or the InfraRed spectrum can be used to acquire images at specific wavelengths. Promising results have been obtained in these cases and some of those works surveyed are highlighted in Table \ref{subbands}.  

A database needs to be acquired consisting of images of the biometric trait for every individual enrolled in the system. Once the images have been captured by the sensors, the templates are obtained by feature selection and extraction. The next step is to match images or templates that are present in the gallery set against images or templates of the incoming probe via template matching or classification. The next section deals with the acquired databases available for public use.

\begin{table}
\centering
\begin{tabular}{ | l | l | }

\hline
	\textbf{Band} & \textbf{Wavelength} \\ \hline
	Visible & 350-740 $\eta$m \\ 
	Near InfraRed (NIR) & 740-1000 $\eta$m \\ 
	Short-Wave InfraRed (SWIR) & 1-3 $\mu$m \\ 
	Mid-Wave InfraRed (MWIR) & 3 - 5 $\mu$m \\
	Long-Wave InfraRed (LWIR) & 8-14 $\mu$m \\ 
	\hline
\end{tabular}
\caption{The ranges of visible, active \& passive InfraRed, as seen in Figure \ref{pansy}} \label{xxx}
\end{table}

\section{Databases} \label{publicdata}

When a face recognition system which uses spectral sensors is developed, the choice of database plays an important role for testing such a system. If a large database is easily and publicly available to all researchers in the domain, testing, benchmarking and comparing is fairly straightforward. However, for spectral imaging, acquiring such a common database that may help in surging forward in the domain, has been a challenging task. In this section, we consider the most commonly used and publicly available hyperspectral/multispectral datasets to date. 

\subsection{Carnegie Mellon University Hyperspectral Face Dataset (CMU-HSFD)}

\begin{figure}[!htb]
\centering
\includegraphics[scale=0.09]{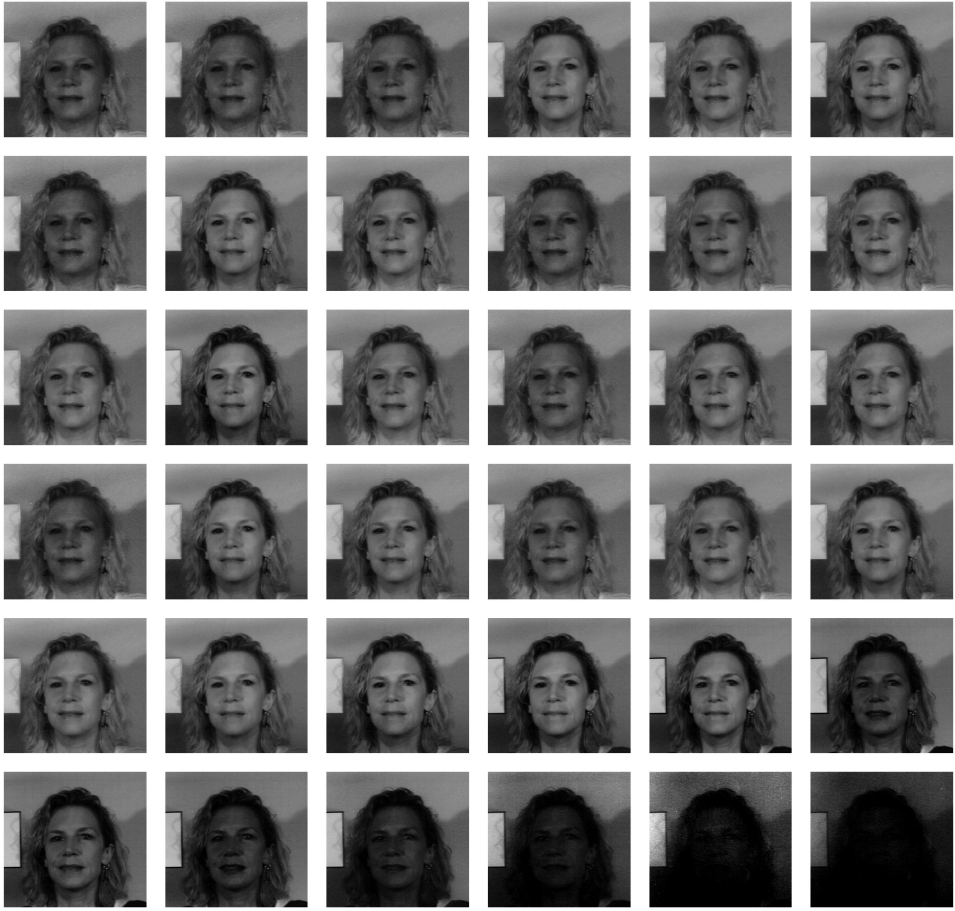}
\caption{CMU-HSFD: Example of a subject's face cube \cite{denes2002hyperspectral}} \label{oo}
\end{figure}

This is a publicly available and widely used hyperspectral face database \cite{denes2002hyperspectral}. The face cubes have been acquired by a spectro-polarimetric camera. The spectral range in which the images have been acquired, is from  450 $\eta$m to 1100 $\eta$m with a step size of 10 $\eta$m. The images have been acquired in multiple sessions from 48 subjects. The illumination conditions have been varied in each session. There is blinking on the part of the subjects owing to the fact that image capture takes several seconds, causing alignment errors. This is one of the most cited datasets for hyperspectral face images and has been used in recent works of Uzair \textit{et al.} \cite{7010906, uzair2013hyperspectral}, Liang \textit{et al.} \cite{liang20153d}, Sharma \textit{et al.} \cite{sharma2016image}, Jing \textit{et al.} \cite{jing2016multi} and Chen \textit{et al.} \cite{chen2016hyperspectral}. In Figure \ref{oo}, we see an example of a subject's cube from the CMU-HSFD database.

\subsection{The Hong Kong Polytechnic University Hyperspectral Face Database (HK PolyU-HSFD)} 

\begin{figure} [!htb]
\centering
\includegraphics[scale=0.7]{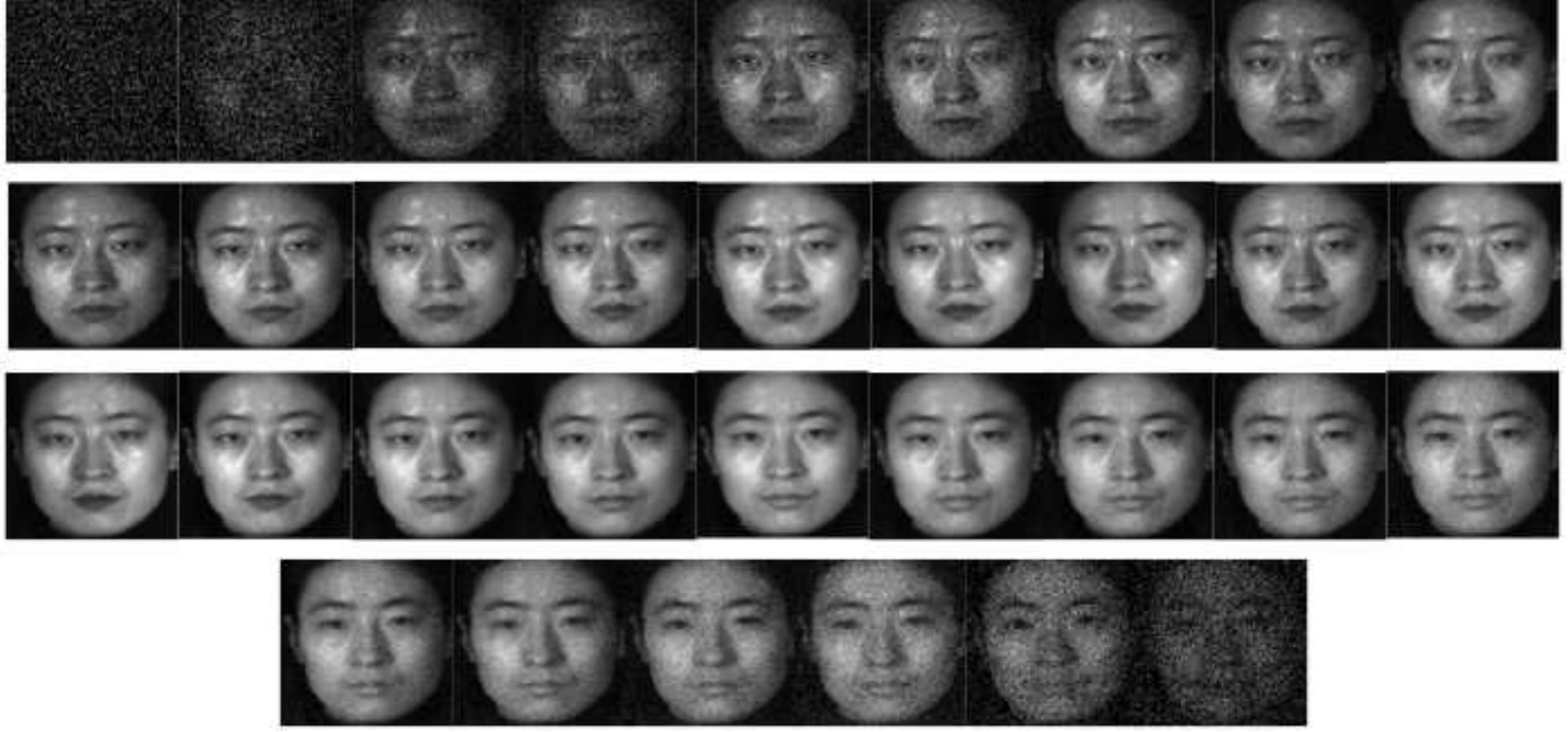}
\caption{HK PolyU-HSFD: Example of a subject's face cube \cite{di2010studies}} \label{cw}
\end{figure}

This is a publicly available dataset consisting of hyperspectral face image cubes \cite{di2010studies}. The database was developed at the Hong Kong Polytechnic University. The face cubes have been acquired using an indoor system made up of CRI’s VariSpec Liquid Crystal Tuneable Filter with a halogen light source. There are 300 hyperspectral face cubes obtained from 25 subjects (17 male subjects and 8 female subjects, age ranging from 21 to 33 years). For each subject, the hyperspectral face cubes were acquired over multiple sessions in an average span of 5 months. In every session, three face cubes were acquired in three views i.e frontal, right and left views with a neutral expression. The spectral range in which the images have been acquired, is from 400 $\eta$m to 720 $\eta$m. As a result of the spectral step size of 10 $\eta$m, 33 bands are obtained. The authors report variations in appearance, for instance, changes of hair style and skin condition, owing to the long timespan in which the images were acquired. This database has also been used recently for testing in their research by Chen \textit{et al.} \cite{chen2016hyperspectral}, Jing \textit{et al.} \cite{jing2016multi}, Uzair \textit{et al.} \cite{uzair2013hyperspectral, 7010906}, Liang \textit{et al.} \cite{liang20153d} and Sharma \textit{et al.} \cite{sharma2016image}. Figure \ref{cw} shows an example of a face cube from the HK PolyU-HSFD.

\subsection{UWA Hyperspectral Face Database (UWA-HSFD)}

\begin{figure}[!htb]
\centering
\includegraphics[scale=0.06]{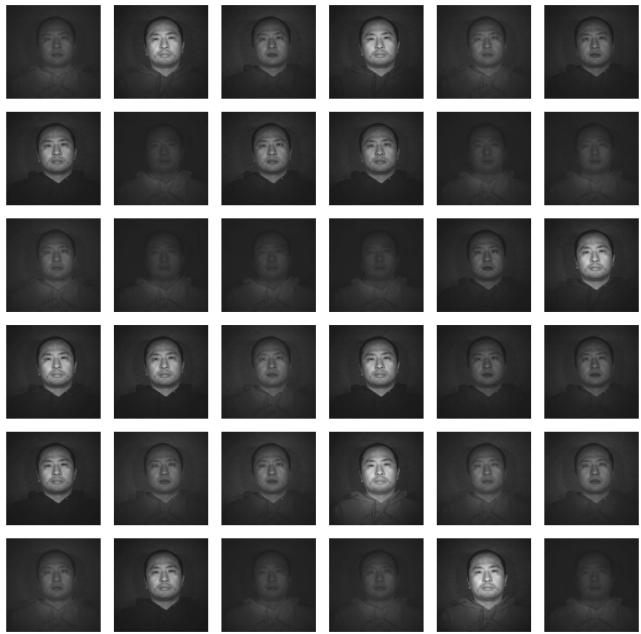}
\caption{UWA-HSFD: Example of a subject's face cube \cite{7010906, uzair2013hyperspectral}} \label{yy}
\end{figure}

The third database that has been made public recently is the UWA Hyperspectral Face Database (UWA-HSFD). This database consists of hyperspectral face image cubes. An indoor imaging system has been used to capture these image cubes. CRI’s VariSpec Liquid Crystal Tuneable Filter has been used, similar to the HK PolyU-HSFD, integrated with a Photon focus camera. The spectral range in which the images have been acquired, is from  400 $\eta$m to 720 $\eta$m with a step size of 10 $\eta$m. As discussed by Uzair \textit{et al.} \cite{7010906, uzair2013hyperspectral}, the camera exposure time was chosen and altered according to the signal-to-noise ratio in different bands. Therefore, as compared to the other two databases, this dataset has the advantage of having lower noise levels. There were 70 subjects for this dataset. Due to eye blinking, there are alignment errors. More details of this dataset can be found in the works of Uzair \textit{et al.} \cite{7010906, uzair2013hyperspectral}. An example of a face cube from the UWA-HSFD database can be seen in Figure \ref{yy}.

\subsection{The Hong Kong Polytechnic University (PolyU) NIR Face Database}

\begin{figure}
\centering
\includegraphics[scale=0.8]{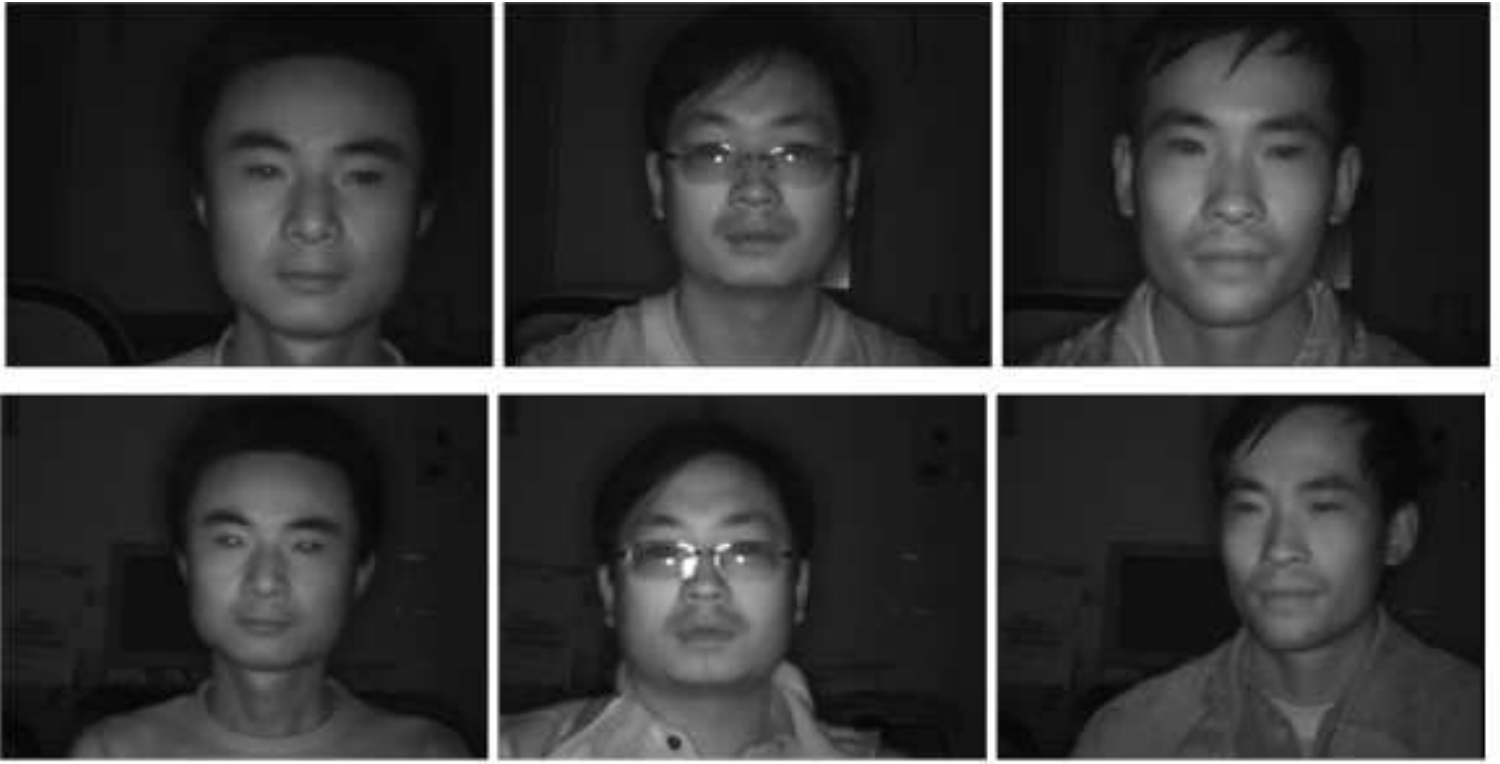}
\caption{PolyU NIR Database: Examples of NIR images of a subject \cite{zhang2010directional}} \label{ty}
\end{figure}

This is a publicly available database of NIR images of test subjects, details of which can be found in the work of Zhang \textit{et al.} \cite{zhang2010directional}. The database was developed at the Hong Kong Polytechnic University. The images were captured from 335 subjects. Originally 350 were captured but the data was only released for 335 subjects due to missing data of the remaining subjects. Frontal images of the subjects were captured initially. Later, pose and facial expression variations were also recorded. Scale variations were also recorded by having the subjects close to or far from the camera. Further to this, images of 15 subjects were captured in two different time intervals with a time difference of two months. For these subjects, 100 images were captured each. A total of 34000 images were captured for the PolyU NIR Face Database. An example of a face cube from the HK PolyU NIR Face Database can be seen in Figure \ref{ty}. 

\subsection{CASIA HFB Database}

The CASIA HFB Database is a publicly available database of heterogeneous face images \cite{li2009hfb}. The heterogeneous face images refer to images of a person in different spectral bands. HFB stands for Heterogeneous Face Biometrics. It consists of images in the visual (VIS) and Near InfraRed (NIR) spectrum. It also contains 3D images of the face. This database was collected and released to study mapping between images captured in different wavelengths. There were a total of 100 subjects, of which 57 were male candidates while 43 were female candidates. For each subject, 4 images were captured in the VIS band and 4 in the NIR band. For the case of 3D images, 2 images were captured each for 92 subjects while 1 image for rest of the 8 subjects. The limitations of this database as per Li \textit{et al.} \cite{6595898} are:

\begin{enumerate}

\item Small number of subjects.

\item Lack of protocols for reproduction of experimental results and evaluation of performance. 

\end{enumerate}  

\begin{figure}
\centering
\includegraphics[scale=0.8]{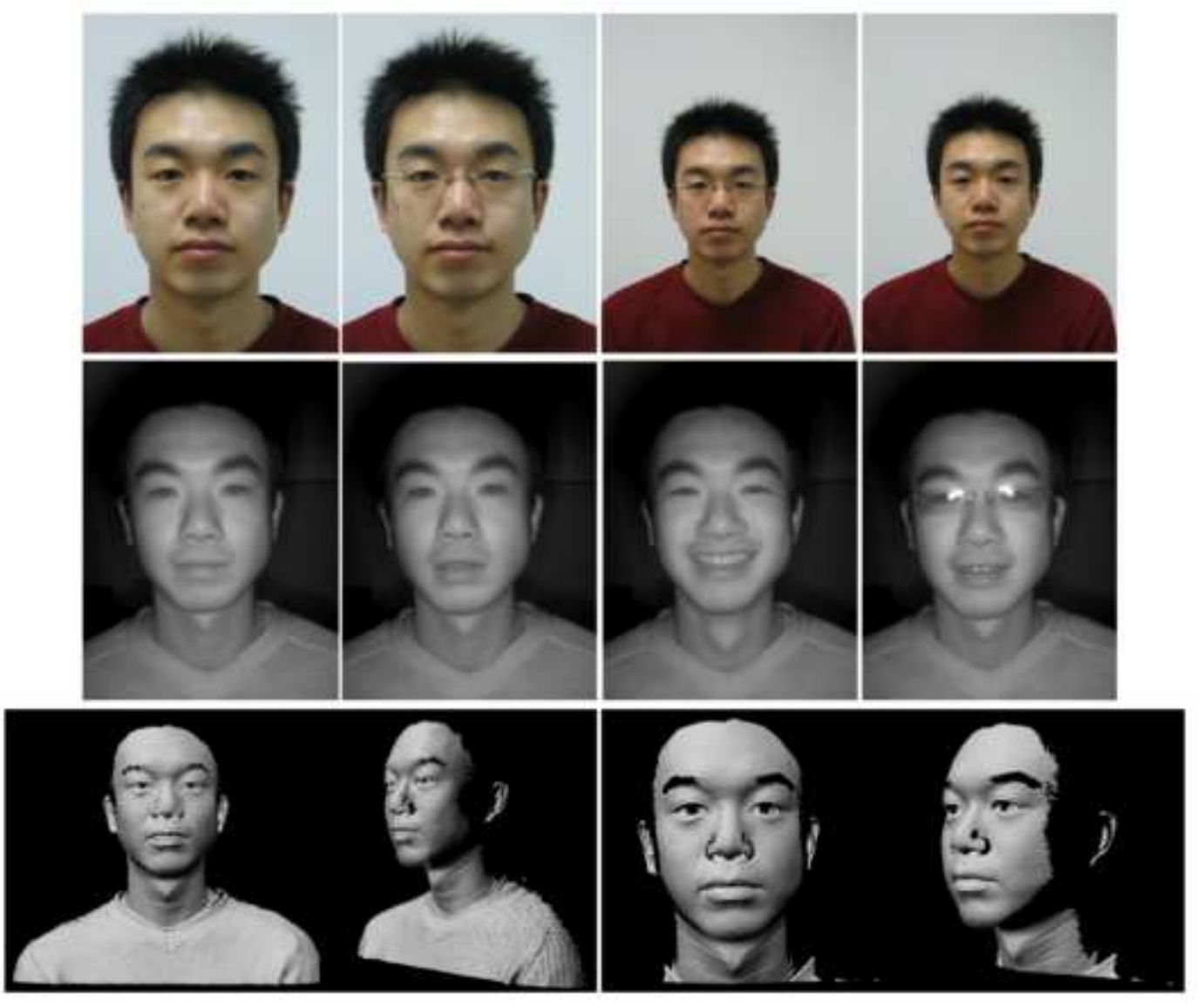}
\caption{CASIA HFB Database: VIS, NIR \& 3D face images \cite{li2009hfb}} \label{pp}
\end{figure}

An example of a face cube from the CASIA HFB Database can be seen in Figure \ref{pp}.

\subsection{CASIA NIR-VIS 2.0 Database}

\begin{figure}
\centering
\includegraphics[scale=1.3]{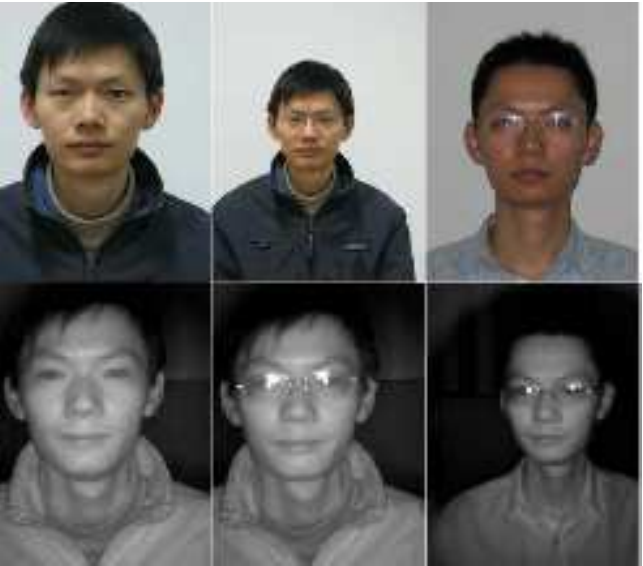}
\caption{CASIA NIR-VIS 2.0 Face Database: VIS \& NIR images \cite{6595898}} \label{cp}
\end{figure}

This is a publicly available database for face recognition research in the NIR-Visible spectrum. This database is an improvement of CASIA HFB Database. The database captures images of 725 subjects. The images have been captured in four sessions using both visible and NIR cameras. This database has diversity in age. Li \textit{et al.} highlighted the protocols for performance evaluation along with some baseline results \cite{6595898}. An example of face images in the VIS and NIR range from the CASIA NIR-VIS 2.0 Face Database can be seen in Figure \ref{cp}.  

\subsection {Limits posed by existing datasets:} 
\begin{figure}
\centering
\includegraphics[scale=0.7]{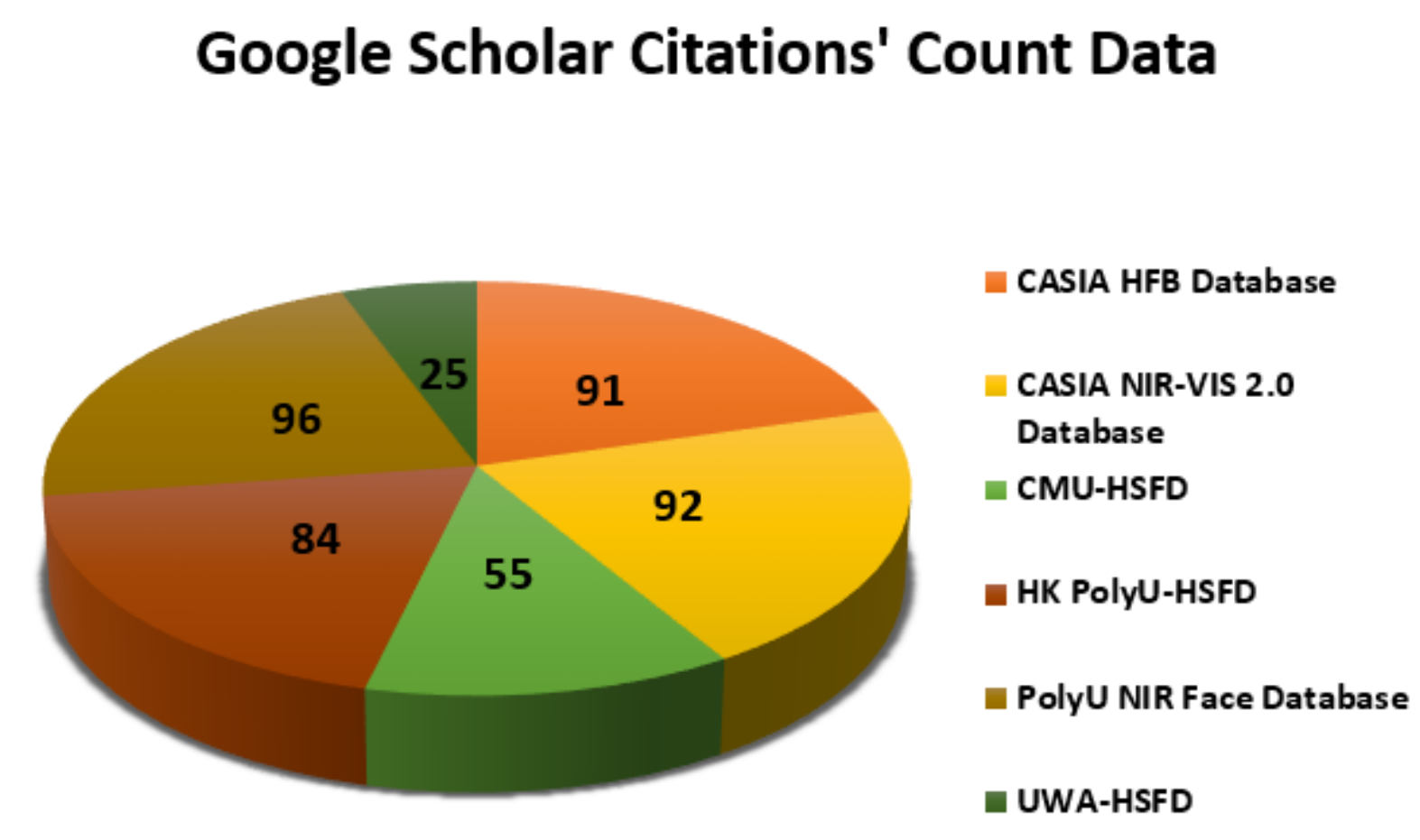}
\caption{The number of times the database has been cited, according to Google Scholar (last accessed: June 2018)} \label{ct}
\end{figure}

These databases are publicly available at the moment however, we note that:
\begin{enumerate}

\item The number of test subjects is not large in any of the datasets.
\item All of these databases are captured in different spectral bands and cannot be cross-evaluated for performance evaluation of different algorithms, due to large modality gap. 
\item There is a large gap in the timeline of the first publication made using the databases. As can be seen in Table \ref{trendtable}, there are major gaps in the years in which these databases were produced. 
\item These databases were not made public in the aforementioned years in Table \ref{trendtable}, rather they were privately used for the first time in the particular year, in a publication. Therefore, it was not possible to reproduce research results for a very long duration. 

\end{enumerate}
Consequently, in Figure \ref{ct} and in Table \ref{trendtable}, the number of citations are based on Google Scholar results of the number of times the database has been cited, at the time of access.

\begin{table*}[!h]
\centering
\begin{tabular}{ | l | c | c | }
\hline
\centering
\textbf{Database} & \textbf{Google Scholar Citation Count} & \textbf{Publication Year} \\ \hline
CMU-HSFD \cite{denes2002hyperspectral} & 55 & 2002 \\ \hline
CASIA HFB Database \cite{li2009hfb} & 91 & 2009 \\ \hline
PolyU NIR Face Database \cite{zhang2010directional} & 96 & 2010 \\ \hline
HK PolyU-HSFD \cite{di2010studies} & 84 & 2010 \\ \hline
CASIA NIR-VIS 2.0 Database \cite{6595898} & 92 & 2013 \\ \hline
UWA-HSFD \cite{7010906, uzair2013hyperspectral} & 25 & 2013 \\ \hline

\end{tabular}
\caption{Database used for the first time in a publication in the corresponding year (last accessed: June 2018)} \label{trendtable}
\end{table*}

We have presented the most widely used databases. A comprehensive list of other IR databases is also presented by Ghiass \textit{et al.} in their survey \cite{ghiass2014infrared}. These databases are available on request. Next section deals with a quantitative analysis of methods applied simultaneously to these datasets.

\onecolumn 
\begin{longtable}{ | p{0.6cm} | p{2cm} | p{9cm} | p{2cm} | }
%\centering
%\begin{tabular}
\hline

\textbf{SN} & \textbf{Symbol} & \textbf{Description} & \textbf{Reference} \\ \hline

1 & AdaBoost & Adaptive Boosting & \cite{freund1996experiments} \\ \hline
2 & AHISD &  Affine Hull-based Image Set Distance & \cite{cevikalp2010face} \\ \hline
3 & BIC & Bayesian Intra-personal/Extra-personal Classifier & \cite{teixeira2003bayesian} \\ \hline
4 & CCA & Canonical Correlation Analysis & \cite{thompson2005canonical} \\ \hline
5 & CDL & Covariance Discriminative Learning & \cite{wang2012covariance} \\ \hline
6 & CHISD & Convex Hull based ImageSet Distance & \cite{cevikalp2010face} \\ \hline
7 & CMLD & Composite Multi-Lobe Descriptor & \cite{doi:10.1117/12.2176462} \\ \hline
8 & CRC & Collaborative Representation Classifier & \cite{zhang2011sparse} \\ \hline
9 & DCC & Discriminant Canonical Correlation Analysis & \cite{kim2007discriminative} \\ \hline
10 & DCT & Discrete Cosine Transform & \cite{uzair2013hyperspectral} \\ \hline
11 & DOG & Difference of Gaussian & \cite{wang2012improved} \\ \hline
12 & DRDWTF & Decimated Redundant Discrete Wavelet Transform and Fisherface & \cite{li2012face} \\ \hline
13 & DSIFT-FVs & Dense SIFT- Fisher Vectors & \cite{mahendran2015understanding, sharma2016image, sharma2016hyperspectral} \\ \hline
14 & DWT & Discrete Wavelet Transform & \cite{shensa1992discrete} \\ \hline
15 & DWTSRDA & Discrete Wavelet Transform + Spectral Regression Discriminant Analysis & \cite{farokhi2014near} \\ \hline
16 & FDA & Fisher Discriminant Analysis & \cite{mika1999fisher} \\ \hline
17 & FPLBP & Four-Patch LBP & \cite{wolf2008descriptor} \\ \hline
18 & FR & Face Recognition & \\ \hline
19 & GDBC & Gabor Wavelet + DBC & \cite{zhang2010directional} \\ \hline
20 & GFC & Gabor Fisher Classifier & \cite{liu2002gabor} \\ \hline
21 & GLBP & Generalized Local Binary Pattern & \cite{guo2008local} \\ \hline
22 & GM-GS-DSDL & multi-view Discriminative and Structured Dictionary Learning with Group Sparsity and Graph Model & \cite{gao2015multi} \\ \hline
23 & GOM & Gabor Ordinal Measures & \cite{chai2014gabor} \\ \hline
24 & GPS & Gabor PCA + Spectral & \cite{wang2014expression} \\ \hline
25 & HGPP & Histogram of Gabor Phase Patterns & \cite{zhang2007histogram} \\ \hline
26 & HK & Hermite Kernels & \cite{farokhi2014near} \\ \hline
27 & HOG & Histogram of Oriented Gradients & \cite{dalal2005histograms} \\ \hline
28 & ICA & Independent Component Analysis & \cite{lee1998independent} \\ \hline
29 & KPCA & Kernel Principal Component Analysis  & \cite{scholkopf1998nonlinear} \\ \hline
30 & LBP & Local Binary Pattern & \cite{ojala1996comparative, ahonen2006face} \\ \hline
31 & LBPF & Local Binary Pattern + Fisherface & \cite{li2007illumination} \\ \hline
32 & LBPL & Local Binary Pattern + Linear Discriminant Analysis & \cite{li2007illumination} \\ \hline
33 & LCVBP &  Local Color Vector Binary Pattern & \cite{lee2011local} \\ \hline
34 & LDA & Linear Discriminant Analysis & \cite{izenman2013linear} \\ \hline
35 & LDP & Local Derivative Pattern & \cite{liang20153d} \\ \hline
36 & LFA & Local Feature Analysis & \cite{penev1996local} \\ \hline
37 & LGBPHS & Local Gabor binary pattern histogram sequence & \cite{zhang20052d} \\ \hline
38 & LOG & Laplacian of Gaussian & \cite{sotak1989laplacian} \\ \hline
39 & LTP & Local Ternary Pattern & \cite{tan2010enhanced} \\ \hline
40 & LVSR-DL & Learning View-invariant Sparse Representations via Dictionary Learning & \cite{zheng2013learning} \\ \hline
41 & MDA & Manifold Discriminant Analysis & \cite{wang2009manifold} \\ \hline 
42 & MLSDL & Multi spectral Low rank Structured Dictionary Learning & \cite{jing2016multi} \\ \hline

43 & MMD & Manifold-Manifold Distance & \cite{wang2008manifold} \\ \hline          
44 & NN & Nearest Neighbour & \cite{beis1997shape} \\ \hline
45 & NRIQA & Non-Reference based Image Quality Assessment & \cite{NARANG201526} \\ \hline
46 & OLPP & Orthogonal Locality Preserving Projection & \cite{cai2006orthogonal} \\ \hline
47 & PCA & Principle Components Analysis & \cite{abdi2010principal, wold1987principal} \\ \hline
48 & PHOG & Pyramid HOG & \cite{bosch2007representing, dalal2005histograms} \\ \hline
49 & PZMRBF & Pseudo Zernike Moments+Radial Basis Function Neural Network & \cite{haddadnia2003efficient} \\ \hline
50 & RBF & Radial Basis Function & \cite{musavi1992original} \\ \hline
51 & RDLF & Rank-based Decision Level Fusion  & \cite{chang2010fusing} \\ \hline
52 & RIQA & Reference based Image Quality Assessment & \cite{NARANG201526} \\ \hline
53 & SANP & Sparse Approximated Nearest Points & \cite{hu2012face} \\ \hline
54 & SIFT & Scale-Invariant Feature Transform & \cite{lowe2004distinctive} \\ \hline
55 & SRC & Sparse  Representation  based  Classification & \cite{wright2008robust} \\ \hline
56 & STE & Sparse Tensor Embedding & \cite{zhao2014sparse} \\ \hline
57 & SURF & Speeded Up Robust Features & \cite{bay2006surf} \\ \hline
58 & SVM & Support Vector Machine & \cite{hearst1998support} \\ \hline
59 & SWLD & Simplified Weber Local Descriptor & \cite{chen2010wld, gong2011face} \\ \hline
60 & TPLBP & Three-Patch LBP & \cite{wolf2008descriptor} \\ \hline
61 & UDWTSRDA & Undecimated Discrete Wavelet Transform + Spectral Regression Discriminant Analysis & \cite{farokhi2015near} \\ \hline
62 & UMD$^2$L & Uncorrected Multi-view Discrimination Dictionary Learning & \cite{jing2014uncorrelated} \\ \hline
63 & WLD & Weber Linear Descriptor & \cite{chen2010wld} \\ \hline
64 & WS & Wavelet Scattering & \cite{chang2012applying} \\ \hline
65 & ZM & Zernike Moments & \cite{farokhi2014near} \\ \hline
66 & ZMHK & Zernike Moments and Hermite Kernels & \cite{farokhi2015near} \\ \hline
67 & ZMSRDA & Zernike Moments + Spectral Regression Discriminant Analysis & \cite{farokhi2015near} \\ \hline
68 & ZMUDWT & Zernike Moments and Undecimated Discrete Wavelet Transform & \cite{farokhi2014near} \\ \hline

%\end{tabular}
\caption{Table of Acronyms} \label{acronym}
\end{longtable}

\section{Quantitative Analysis of Methods on Datasets} \label{quant}

This section deals with a quantitative analysis of different methods on the databases cited in the previous section. This section first, compares methods applied to all three hyperspectral databases. It then accounts for results obtained from different methods applied to all public NIR databases. Works consisting of methods applied to a single dataset from the previous section is out of scope of this review. We have included works that simultaneously provide comparisons of methods on either all three hyperspectral datasets or all NIR datasets, listed in the previous section. 

Liang \textit{et al.} proposed to use a method called 3D Local Derivative Pattern (LDP) \cite{liang20153d}. This method is a 3D texture descriptor which analyzes micro-patterns and maps them to binary. Liang \textit{et al.} compare the results of their method against various other methods and find that their method outperforms the rest. Table \ref{quantsdatasets} shows recognition rates of the various methods. They have used two datasets for experiment, the HK-PolyU Hyperspectral Face Database and the CMU Hyperspectral Face Dataset. Uzair \textit{et al.} \cite{uzair2013hyperspectral} use 3D-Discrete Cosine Transform (DCT) for feature extraction and both, Sparse Representation based Classification (SRC) and Partial Least Squares (PLS) regression for classification on all three hyperspectral datasets. Recognition rates obtained by the proposed technique and those by other hyperspectral algorithms are shown in table \ref{quantsdatasets}. Further, Uzair \textit{et al.} \cite{7010906} also developed another method which uses spatiospectral covariance for band fusion. The technique extracts a single fused, most discriminatory image from the hyperspectral cube. In table \ref{quantsdatasets}, we have highlighted the results of their technique against image-set classification algorithms and Grayscale and RGB FR algorithms. Jing \textit{et al.} \cite{jing2016multi} use multi-view dictionary learning method for classification and apply it to all three datasets with different levels of corruption added to the pixels. We note the results for 0 pixel corruptions in Table \ref{quantsdatasets}. 

Farokhi \textit{et al.} \cite{farokhi2014near} combine both local and global features: Undecimated Discrete Wavelet Transform, denoted as UDWTSRDA (local features) and Zernike moments, denoted as ZMSRDA (global features)These are combined to give Decision Fusion, denoted by DF.  They apply the proposed method to both CASIA NIR database and PolyU-NIR Face database. Table \ref{quantsNIR} shows a comparison of results obtained from the proposed method against other methods, under different facial expressions. Further, Farokhi \textit{et al.} \cite{farokhi2015near} proposed a technique based on Zernike Moments and Hermite Kernels to cater for variations in facial expressions or presence of occlusion. The results produced from application of this technique compared to other techniques are presented in Table \ref{quantsNIR}.

\begin{table}
\centering

\begin{tabular}{lllcccccc}
\toprule
\textbf{Author(s), Year} &  &\textbf{Method}  & \multicolumn{4}{c}{\textbf{Values}} \\
\cmidrule{4-6}
 & & & CMU-HSFD & HK PolyU-HSFD & UWA-HSFD &  \\
\midrule
Uzair \textit{et al.} \cite{uzair2013hyperspectral}, 2013 & & Spectral Signature & \SI{38.18 \pm 1.89} & \SI{24.63 \pm 3.87} & \SI{40.52 \pm 1.08} & \\ 
       &    & Spectral Angle & \SI{38.16 \pm 1.89} & \SI{25.49 \pm 4.36} & \SI{37.95 \pm 4.15} & \\

       &    & Spectral Eigenface & \SI{84.54 \pm 3.78} & \SI{70.30 \pm 3.61} & \SI{91.51 \pm 3.07} & \\
		   &    & 2D PCA & \SI{72.10 \pm 5.41} & \SI{71.11 \pm 3.16} & \SI{83.85 \pm 2.42} & \\
		   &    & 3D Gabor Wavelets & \SI{91.67 \pm 2.86} & \SI{90.19 \pm 2.09} & \SI{91.50 \pm 3.07} & \\
			
			 &    & 2D DCT+SRC & \SI{97.44 \pm 1.24} & \SI{75.86 \pm 2.92} & \SI{97.00 \pm 1.29} & \\
		   &    & 2D DCT+PLS & \SI{97.78 \pm 1.28} & \SI{91.43 \pm 2.10} & \SI{97.25 \pm 1.87} & \\
			 &    & 3D DCT+SRC & \SI{98.10 \pm 0.69} & \SI{87.02\pm 1.72} & \SI{98.00 \pm 1.84} & \\
		   &    & 3D DCT+PLS & \SI{99.00 \pm 0.85} & \SI{93.00 \pm 2.27} & \SI{98.00 \pm 1.39} & \\

\midrule
Liang \textit{et al.} \cite{liang20153d}, 2015 & & Spectral Feature &\SI{38.18 \pm 1.89} & \SI{45.35 \pm 3.87} & - & \\ 
       &    & Spectral Eigenface & \SI{84.54 \pm 3.78} & \SI{70.33 \pm 3.61} & - & \\
		   &    & 2D PCA & \SI{72.10 \pm 5.41} & \SI{71.00 \pm 3.16} & - & \\
		   &    & 3D DCT & \SI{88.65 \pm 2.34} & \SI{84.00 \pm 3.35} & - & \\
			 &    & 3D LBP & \SI{92.16 \pm 3.52} & \SI{88.8 \pm 1.79} & - & \\
		   &    & 3D Gabor Wavelet & \SI{92.20 \pm 2.46} & \SI{90.00 \pm 2.83} & - & \\
			 &    & 3D LDP & \SI{94.83 \pm 2.62} & \SI{95.33 \pm 1.63} & - & \\
		   &    & 2D LDP & \SI{90.27 \pm 3.51} & \SI{86.25 \pm 9.62} & - & \\
			
\midrule
Uzair \textit{et al.} \cite{7010906}, 2015 &  & DCC & \SI{87.5 \pm 3.04} & \SI{76.0 \pm 3.72} & \SI{91.5 \pm 3.58} & \\

		   &    & MMD & \SI{90.0 \pm 2.78} & \SI{83.8 \pm 4.16} & \SI{82.8 \pm 4.16} & \\
		   &    & MDA & \SI{90.6 \pm 2.87} & \SI{87.9 \pm 2.84} & \SI{91.0 \pm 3.16} & \\
			 &    & AHISD & \SI{90.6 \pm 2.82} & \SI{89.9 \pm 2.98} & \SI{92.5 \pm 2.04} & \\
		   &    & CHISD & \SI{91.1 \pm 2.31} & \SI{90.3 \pm 1.38} & \SI{92.5 \pm 2.36} & \\
			 &    & SANP & \SI{90.9 \pm 2.77} & \SI{90.5 \pm 1.47} & \SI{92.5 \pm 2.19} & \\
		   &    & CDL & \SI{92.7 \pm 2.31} & \SI{89.3 \pm 2.56} & \SI{93.1 \pm 1.50} & \\
						
       &    & Eigenfaces & \SI{82.6 \pm 4.12} & \SI{76.5 \pm 4.48} & \SI{80.5 \pm 2.32} & \\
		   &    & Fisherfaces & \SI{93.6 \pm 1.87} & \SI{82.2 \pm 2.46} & \SI{96.0 \pm 1.32} & \\
		   &    & LBP & \SI{96.6 \pm 1.67} & \SI{90.0 \pm 2.42} & \SI{96.2 \pm 1.05} & \\
			 &    & SRC & \SI{91.0 \pm 2.08} & \SI{85.6 \pm 2.49} & \SI{96.2 \pm 1.50} & \\
		   &    & CRC & \SI{93.8 \pm 1.81} & \SI{86.1 \pm 2.37} & \SI{96.2 \pm 1.00} & \\
			 &    & LCVBP+RLDA & \SI{87.3 \pm 2.62} & \SI{80.3 \pm 6.08} & \SI{97.0 \pm 1.45} & \\
		   &    & Band fusion+PLS & \SI{99.1 \pm 0.57} & \SI{95.2 \pm 1.60} & \SI{98.2 \pm 1.21} & \\

\midrule
Jing \textit{et al.} \cite{jing2016multi}, 2016   &    & RDLF & \SI{83.97 \pm 4.21} & \SI{80.77 \pm 3.64} & \SI{88.01 \pm 2.61} & \\

		   &    & STE & \SI{85.16 \pm 4.85} & \SI{81.15 \pm 3.91} & \SI{89.85 \pm 2.84} & \\
			 &    & GPS & \SI{88.29 \pm 2.02} & \SI{84.07 \pm 1.20} & \SI{93.17 \pm 1.61} & \\
		   &    & LVSR-DL & \SI{93.73 \pm 1.30} & \SI{85.33 \pm 1.40} & \SI{96.60 \pm 1.32} & \\
			 &    & UMD$^2$L & \SI{94.54 \pm 1.91} & \SI{85.79 \pm 2.02} & \SI{97.04 \pm 1.57} & \\
		   &    & GM-GS-DSDL & \SI{91.43 \pm 2.54} & \SI{84.25 \pm 2.61} & \SI{94.97 \pm 1.80} & \\
			 &    & MLSDL & \SI{95.88 \pm 1.17} & \SI{86.24 \pm 1.84} & \SI{98.51 \pm 1.04} & \\

\bottomrule
\end{tabular}
\caption{Quantitative results (average recognition rates with standard deviation in percentage) of different methods on the cited datasets: CMU-HSFD, HK PolyU-HSFD and UWA-HSFD}\label{quantsdatasets}
\end{table}

\begin{table}
\centering

\begin{tabular}{lllcccccc}
\toprule
\textbf{Author(s), Year} &  &\textbf{Method}  & \multicolumn{3}{c}{\textbf{Values}} \\
\cmidrule{4-5}
 & & & CASIA NIR & HK PolyU-NIR &  \\
\midrule
Farokhi \textit{et al.} \cite{farokhi2014near}, 2014 & & LDA & \SI{91.12 \pm 2.03} & \SI{93.25 \pm 1.26} & \\ 
       &    & PCA & \SI{75.45 \pm 3.82} & \SI{91.66 \pm 1.48}  & \\

       &    & KPCA & \SI{83.20 \pm 2.21} & \SI{83.20 \pm 2.21}  & \\
		   &    & PZMRBF & \SI{86.76 \pm 1.24} & \SI{88.66 \pm 2.01}  & \\
		   &    & OLPP & \SI{93.16 \pm 1.01} & \SI{95.29 \pm 1.48}  & \\
			
			 &    & LBPF & \SI{94.62 \pm 1.30} & \SI{94.58 \pm 1.28}  & \\
		   &    & DRDWTF & \SI{92.54 \pm 1.13} & \SI{93.70 \pm 1.76}  & \\
			 &    & DWTSRDA & \SI{91.04 \pm 1.29} & \SI{93.22 \pm 1.04} & \\
		   &    & ZMSRDA & \SI{92.59 \pm 1.32} & \SI{93.91 \pm 1.11}  & \\
			 &    & UDWTSRDA & \SI{93.41 \pm 0.71} & \SI{95.83 \pm 0.90}  & \\
			 &    & DF & \SI{96.58 \pm 0.74} & \SI{97.58 \pm 0.71}  & \\

\midrule
Farokhi \textit{et al.} \cite{farokhi2015near}, 2015 & & GFC & \SI{80.26 \pm 3.61} & \SI{76.79 \pm 2.31} & \\ 
       &    & LBPL & \SI{86.75 \pm 2.93} & \SI{81.57 \pm 4.47}  & \\

       &    & GDBC & \SI{80.14 \pm 4.39} & \SI{75.62 \pm 3.56}  & \\
		   &    & WS & \SI{82.79 \pm 2.31} & \SI{80.16 \pm 3.64}  & \\
		   &    & ZMUDWT & \SI{90.25 \pm 1.21} & \SI{85.01 \pm 1.78}  & \\
			
			 &    & ZM & \SI{78.29 \pm 3.32} & \SI{76.14 \pm 2.18}  & \\
		   &    & HK & \SI{87.86 \pm 2.46} & \SI{84.26 \pm 2.53}  & \\
			 &    & ZMHK & \SI{91.47 \pm 2.13} & \SI{87.22 \pm 2.11} & \\

\bottomrule
\end{tabular}
\caption{Quantitative results (average recognition rates with standard deviation in percentage) of different methods on the cited NIR datasets: CASIA NIR and HK PolyU-NIR}\label{quantsNIR}
\end{table}

\section{Spectral Matching} \label{spectralmatch}

Once the database has been obtained for spectral images, test images or probes can be matched or classified.   
In this process, probe set images (testing images) in one modality are matched or classified against gallery set images (database images or training images) obtained in the same modality or a different modality, stored in the gallery set (database). Interspectral matching refers to the case where images in the gallery set and the probe set are captured in different spectral bands. In section \ref{intermatch}, we proceed with some of the works in inter-spectral matching in either active IR domain, passive IR domain or matching of sub-bands. Intra-spectral matching refers to the case where images in the gallery set and the probe set are captured in the same spectral band. Section \ref{intramatch} deals with works in intra-spectral matching, in either active IR domain, passive IR domain or matching of sub-bands.  

Some factors affect the identification of individuals by biometric systems, for which high-confidence matching algorithms need to be developed. Some of those factors are as under:

\begin{enumerate}

\item The images in the gallery set and the probe set may or may not be in same spectral bands. This will consequently require use of case-by-case matching algorithms. Refer to sections \ref{intermatch} and \ref{intramatch} for relevant works in matching processes.  
\item The captured images could be in controlled or uncontrolled conditions (changes in illumination direction and stand-off distances) which could in turn, affect the matching results \cite{6284307}.
\item Facial expressions, cosmetics, pose variations and facial disguise may affect matching results \cite{steiner2016reliable, Ramachandra:2017:PAD:3058791.3038924, raghavendra2017vulnerability, raghavendra2017face}. 

\end{enumerate}

There has been extensive work in matching InfraRed bands to visible band. Also, some spectral sub-bands in the visible \& IR spectrum have given better recognition rates as compared to other sub-bands, due to which sub-band selection has also been applied by Bouchech \textit{at al.} \cite{Bouchech:2015:KSA:2821861.2821895} and Chang \textit{et al.} \cite{chang2009improving}. Since the reflection dominant bands of InfraRed (active) are closer to the visible range, there have been good inter-spectral matching results (between visible and NIR or SWIR). But as we move further from the visible range in the electromagnetic spectrum, the emission dominant bands (passive or thermal) do not provide good matching results since there is a large modality gap between these two modalities. Hence, exploiting the merits of the thermal band has proved to be a challenge for face recognition.  

We highlight our investigation of the research work, where the gallery set and probe set images are either in the same or in different spectral bands. For different spectral bands, inter-spectral matching or cross-spectral matching poses challenges, as can be seen in section \ref{intermatch}. For instance, depending on the application area, we would want to have SWIR face images which can be used for night time surveillance, as the incoming image, and be able to successfully match it against VIS images in the gallery set. But inter-spectral matching is a very challenging task. 

\subsection{Inter-spectral Matching} \label{intermatch}

In this section, we highlight most novel works available in literature, for the case of inter-spectral matching. Interspectral matching deals with the case where images in the gallery set (database images/training images) are captured in a different spectral band or sub-band compared to the images captured in the probe set (testing images). 

Bourlai \textit{et al.} \cite{10.1117/12.918899} used cross-matching to match MWIR images to visible face images; however, the identification performance did not yield encouraging results. This is due to the fact that, in thermal-to-visible face recognition there is a large modality gap. For features, they used LBP, LTP, PHOG, SIFT, TPLBP and FPLBP. The best identification rate was obtained by TPLBP at 53.9\%.

Short \textit{et al.} \cite{Short:15} reported improvements in cross-modal (LWIR-Visible) face recognition taking polarimetric imaging into account for the LWIR spectrum. HOG was used for feature extraction and high identification rates were reported for their proposed method.

Zhu \textit{et al.} \cite{6710158} matched Visual against Near InfraRed  (VIS-NIR) images of the face. The probe set in their work, consisted of NIR images of the subjects. While those that made up the gallery set, in the enrolment phase, consisted of VIS images. They used both LBP and HOG and reported improved matching results with their proposed method. 

Goswami \textit{et al.} \cite{goswami2011evaluation} matched Near InfraRed images to visible images. They used LBP operator for feature extraction. According to them, their visible to NIR matching outperformed their NIR to visible matching.

Klare \textit{et al.} \cite{5597000} matched NIR images in the probe set against visible images in the gallery set. They used both LBP and HOG. They reported an accuracy of 88.8\% with HOG, higher than what they obtained from LBP. 
 
Kalka \textit{et al.} \cite{6117586} matched SWIR images captured at 1550 $\eta$m (probe set) against visible images (gallery set). However, these images were acquired under variable conditions. In the first case, the images were collected in controlled indoor environment. In the second, images captured in a semi-controlled environment at stand-off distances of 50 m and 106 m. In the third, images were captured during night and day, at stand-off distances of 60-400m. They used both LBP and LTP. For the first case of controlled conditions, they obtained a very high identification result. The second accuracy result is also as high as 90\% but for the third case of large stand-off distance, the accuracy rate falls to 79\%. 

Chang \textit{et al.} \cite{4712365} chose few, specific spectral bands to achieve better recognition performance results. Images in specific spectral bands were obtained and matched against visible images. The spectral bands were in the range of 480-720 $\eta$m. A 20 $\eta$m increment was applied. Improved recognition results obtained on specific bands such as at 610 $\eta$m and 640 $\eta$m.

Mendez \textit{et al.} \cite{10.1007/978-3-642-01793-3_34} used LBP for robust results for face recognition where images were captured in the LWIR spectrum. The authors highlighted the fact that variations in illumination limit the performance of a facial recognition system based on visible spectrum. Images captured in different modalities such as Infrared, are more prone to these illumination changes. A problem associated with LWIR images is the presence of non-uniform noise from external noise sources. An identification rate of 97.3\% was achieved. 

Nicolo \textit{et al.} \cite{6272346} performed matching between SWIR and Visible images. They used Gabor filters \cite{daugman1988complete, jain1997object, lee1996image} resulting in magnitude and phase of images, to which they applied three other operators: SWLD, LBP and GLBP, to extract geometric distributions from face. The matcher based on phase patterns encoded by GLBP descriptor provided the lowest matching performance. The magnitude patterns encoded by SWLD and LBP descriptors both provided significantly higher performance compared to GLBP. 

\subsection{Intra-spectral Matching} \label{intramatch}

In this section, we highlight some of the important works, available in literature used to build the ground for comparison of results from inter-spectral matching with those of intra-spectral matching. Intra-spectral matching deals with the case where images in the gallery set (database images/training images) are captured in the same spectral band or sub-band as the images captured in the probe set (testing images). 

Nicolo \textit{et al.} \cite{6272346} performed intra-spectral matching. For intra-spectral matching, the deployed matcher (as mentioned in Section \ref{intermatch}) worked well for both color and SWIR images when compared with commercial Faceit G8. This work incorporated inter-spectral matching as well. 

Chen \textit{et al.} \cite{chen2005ir} used thermal imagery for intra-spectral matching. They first experimented with PCA based visible imagery and compared the results with commercial Faceit G8. Further to it, they experimented with PCA based thermal IR imagery and compared its results with commercial Faceit G8. PCA-based recognition using visible-light images performed better than PCA-based recognition using IR images, FaceIt-based recognition using visible-light images outperformed both PCA based recognition on visible-light images and PCA-based recognition on IR images. 

Bourlai \textit{et al.} \cite{10.1117/12.918899} used intra-spectral matching to match MWIR images to MWIR images. They obtained 97\% identification rate with LBP. With encouraging results, they were able to conclude that the use of MWIR is advantageous in certain operating conditions. At night or in the dark, the MWIR imagery could give identification performance equivalent to that obtained by visible-visible matching. 

In research, intra-spectral matching has been used to establish baseline results against which evaluations can be performed for the results obtained by inter-spectral matching. The accuracy rates for face recognition suffer in the case of inter-spectral matching when they are compared with the rates of intra-spectral matching \cite{chen2005ir}. The crux of the work related to spectral imaging deals with addressing the challenges posed by inter-spectral matching, rather than intra-spectral matching, which is only used to establish baseline conditions on which results in inter-spectral matching are built upon \cite{6284307}.  

\section{Literature Survey in Active InfraRed \& Passive InfraRed for Face Recognition} \label{apIR}

\begin{figure}
\centering
\includegraphics[scale=0.75]{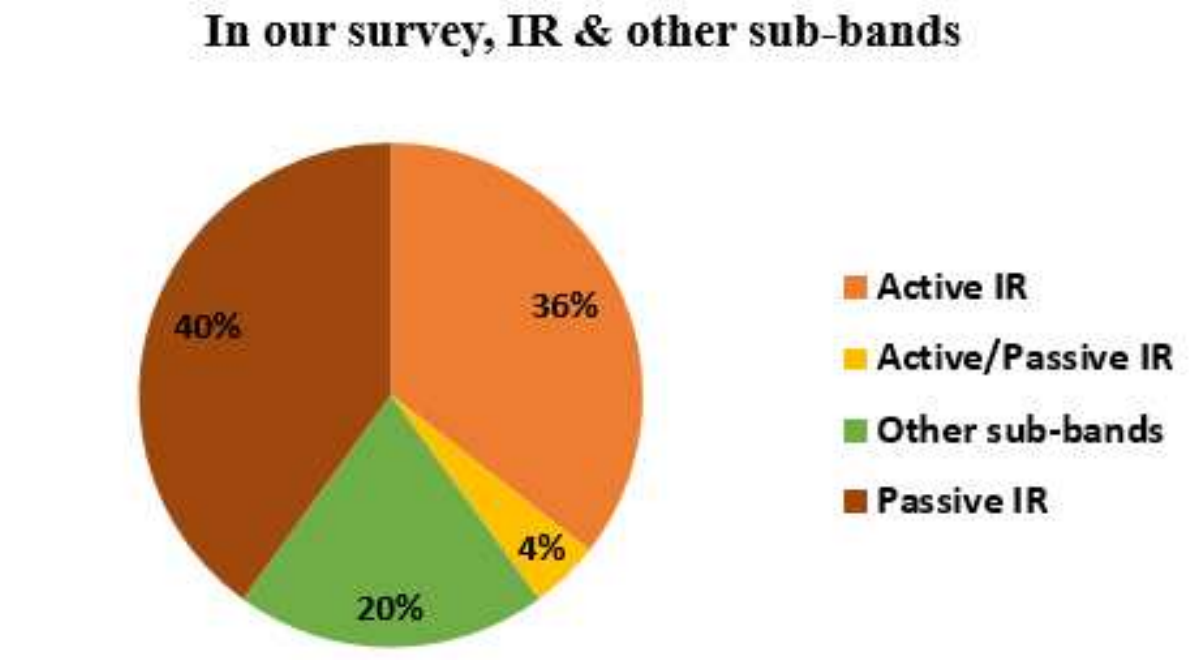}
\caption{Distribution of the research work in Active InfraRed, Passive InfraRed \& other sub-bands used for this survey} \label{yellow}
\end{figure}

The previous section dealt with the types of matching processes in spectral imaging. This section builds upon the previous section by taking into account, the work of the research community in matching images through different techniques obtained in either active IR or passive IR against those in the visible spectrum. We highlight some of the important cited works in literature in the domain of active InfraRed imaging and passive InfraRed imaging for face recognition. Figure \ref{yellow} shows the distribution of the research work that has been used for this survey in either passive or active IR. 36\% of the papers are based in the active IR domain while 40\% of the papers we used are in the passive IR domain, which needs more attention by the research community due to the difference in phenomenology of thermal IR bands compared to visible band. While 20\% are in different sub-bands of the electromagnetic spectrum, the remaining 4\% are based in both IR domains. 

As discussed in section \ref{promisingIR}, active IR band is reflection dominant. It is further divided into Near InfraRed (NIR) and Short Wave InfraRed (SWIR). In Table \ref{ActiveIR}, we have tabulated the most novel literature work in the domain of active InfraRed. 

In Table \ref{PassiveIR}, we tabulate the most novel literature work for passive InfraRed imaging or thermal imaging of the face. As discussed in section \ref{promisingIR}, passive IR band is emission dominant. It is further divided into Mid-Wave InfraRed (MWIR) and Long Wave InfraRed (LWIR). Inter-spectral matching of passive InfraRed images with images of the visible spectrum poses a challenge due to large difference in the phenomenology of the two \cite{10.1117/12.918899, klare2013heterogeneous} and therefore, requires development of robust techniques for successful matching of the two modalities. 

Table \ref{activepassive} takes into account, the specific case where the authors have catered for both cases: active and passive IR in their work. Special cases have been presented in Table \ref{subbands}, where a survey has been completed to account for the cases where researchers used several spectral sub-bands to gain facial information only in those specific sub-bands and compared the performance of their techniques when matching is performed between face images in the visible spectrum and those obtained in the sub-bands. There are various acronyms throughout these tables. These acronyms are given with their references in Table \ref{acronym}.

\onecolumn 
\begin{landscape}
\begin{longtable}{ | p{1cm} | p{4.5cm} | p{3cm} | p{5cm} | p{8cm} | }
%\centering
%\begin{tabular}
\hline

\textbf{SN} & \textbf{Author, Year} & \textbf{Modality type} & \textbf{Method(s)} & \textbf{Brief Description} \\ \hline

\endhead
\multicolumn{3}{@{}l}{\ldots \textit{continued on next page}}

\endfoot
\endlastfoot

1 & Zou \textit{et al.}, 2005 \cite{zou2005face} & Active & LDA subspace produced from PCA projections.
RBF, SVM, AdaBoost \& NN classifiers used. & Addressed the problem of illumination by use of NIR. Used different classifiers to analyze performance of each.  \\ \hline
2 & Kang \textit{et al.}, 2006 \cite{4107140} & Active & Eigenface \& Fisherface methods used. & Development of a low cost IR system, achieving high recognition results for full frontal face.  \\ \hline
3 & Li \textit{et al.}, 2007 \cite{Li:2007:IIF:1263142.1263441} & Active & Use of LBP operator with LDA \& AdaBoost. & Development of a robust NIR based system in a cooperative scenario. A method used to combat the issue of reflection on eye glasses.  \\ \hline
4 & Bourlai \textit{et al.}, 2010 \cite{5597717} & Active & PCA with k-NN and commercial tools used (Verilook \& G8). & Inter-spectral matching performed between visible images and SWIR images. Fusion applied for improvement in recognition performance.   \\ \hline
5 & Klare \textit{et al.}, 2010 \cite{5597000} & Active & LBP \& HOG features extracted, NN matching and sparse representation matching used.  & Inter-spectral matching of NIR and visible images by random subspace projections and sparse representation classification.  \\ \hline
6 & Nicolo et al. \textit{et al.}, 2011 \cite{robustnicolo} & Active & Use of three operators: Gabor filters, LBP \& GLBP. & A method for cross-spectral matching of SWIR and visible images for face recognition by encoding magnitude and phase responses of Gabor filters.  \\ \hline
7 & Kalka \textit{et al.}, 2011 \cite{6117586} & Active & LBP, LTP \& commercial software G8 used.  & Inter-spectral matching performed between visible images and SWIR images. Standoff distances changed and performance illustrated for controlled and uncontrolled environments.   \\ \hline
8 & Zuo \textit{et al.}, 2012 \cite{6374578} & Active & Gabor filter, SWLD, LBP, GLBP and I-divergence used. & Inter-spectral matching between SWIR and visible images by filtering images with Gabor filters. Then applying Simplified Weber Local Descriptor, Local Binary Pattern and Generalized Local Binary Pattern to create feature templates.  \\ \hline
9 & Nicolo \textit{et al.}, 2012 \cite{6272346} & Active & SWLD, LBP and GLBP applied to magnitude and phase of filtered images (after application of Gabor filters). & Inter-spectral matching of SWIR against visible images by filtering images with Gabor filters.   \\ \hline
10 & Pan \textit{et al.}, 2013 \cite{1251148} & Active & Mahalanobis distance used. & Use of multiple NIR sub-bands for face recognition in presence of pose and expression variations.   \\ \hline
11 & Lemoff \textit{et al.}, 2014 \cite{lemoff} & Active & Fusion of images. & Reviewed the development of TINDERS active-SWIR imaging system and performed experiments to show automated functions that can identify mobile individuals at different distances.  \\ \hline
12 & Zhu \textit{et al.}, 2014 \cite{6710158} & Active & LBP, HOG \& Log-DoG used.  & Used transductive heterogeneous face matching (THFM) to address the problem of unavailability of corresponding visible images of NIR counterparts.  \\ \hline
13 & Whitelam \textit{et al.}, 2015 \cite{Whitelam:2015:AEL:2827888.2828093} & Active & LDA, PCA \& LBP used. & Used SWIR to  determine the location of eyes accurately. Used a robust approach to combat image blurring and compression.  \\ \hline
14 & Cao \textit{et al.}, 2015 \cite{doi:10.1117/12.2176462} & Active & CMLD proposed and compared against other operators such as LBP \&  GOM. & A new operator CMLD used to perform inter-spectral matching between NIR or SWIR images and visible light images.   \\ \hline
15 & Narang \textit{et al.}, 2015 \cite{NARANG201526} & Active & Feature extraction performed by RIQA \& NRIQA methods. Bayesian and NN classifiers used. & Images captured in multiple sub-bands of SWIR in multiple poses. Automated and weighted quality based score level fusion schemes proposed and experiments performed.   \\ \hline
16 & Lezama \textit{et al.}, 2016 \cite{lezama2017not} & Active & CNN used. & Applied deep learning to NIR dataset.  \\ \hline

%\end{tabular}
\caption{In chronological order, most important works in the active IR domain} \label{ActiveIR}
\end{longtable}
\end{landscape}

%%%%%%%%%%%%%%%%%%
\begin{landscape}
\begin{longtable}{ | p{1cm} | p{4.5cm} | p{3cm} | p{5cm} | p{8cm} | }
%\centering
%\begin{tabular}
\hline

\textbf{SN} & \textbf{Author, Year} & \textbf{Modality type} & \textbf{Method(s)} & \textbf{Brief Description} \\ \hline

\endhead
\multicolumn{3}{@{}l}{\ldots \textit{continued on next page}}

\endfoot
\endlastfoot

1 & Socolinsky \textit{et al.}, 2001 \cite{990519} & Passive & Used two methods for recognition: ARENA \cite{sim2000memory} \& Eigenfaces \cite{turk1991eigenfaces}. & Examined the illumination invariance property of thermal LWIR images for performance in variable illumination conditions.  \\ \hline
2 & Socolinsky \textit{et al.}, 2002 \cite{1047436} & Passive & PCA, LDA, LFA, ICA used for FR. & Different FR methods used to evaluate the performance on thermal \& visible images.  \\ \hline
3 & Socolinsky \textit{et al.}, 2004 \cite{thermalsoco} & Passive & Experiments were performed using PCA, LDA and Equinox algorithms. & Evaluated the performance of thermal imaging, in a real, uncontrolled case, unlike other studies based on controlled conditions for thermal imaging.   \\ \hline
4 & Chang \textit{et al.}, 2006 \cite{1640494} & Passive & - & Collection of a database in thermal band and subbands of the visible spectrum. Fusion methods applied to cater for varying illumination.   \\ \hline
6 & Buddharaju \textit{et al.}, 2007 \cite{Buddharaju:2007:PFR:1263142.1263440} & Passive & Used PCA for FR. & Proposed method for extraction of features from thermal images based on physiological information.  \\ \hline
7 & Mendez \textit{et al.}, 2009 \cite{10.1007/978-3-642-01793-3_34} & Passive & LBP used on LWIR images & Addressed the problem of fixed pattern noise in illumination invariant LWIR imaging with the help of LBP.  \\ \hline
8 & Martinez \textit{et al.}, 2010 \cite{5543605} & Passive & Haar features for detection of eyes and nostrils. & Worked on detection of facial features in 'ghostly' thermal images.  \\ \hline
9 & Hermosilla et al. \textit{et al.}, 2012 \cite{hermosilla2012comparative} & Passive & LBP, WLD, SIFT, SURF \& Gabor jet used. & Face identification in unconstrained environment using thermal imagery. Local Binary Pattern, Weber Linear Descriptors , Gabor Jet Descriptors, Scale-Invariant Feature Transform and Speeded Up Robust Features Descriptors, based methods used for matching.  \\ \hline
10 & Bourlai \textit{et al.}, 2012 \cite{10.1117/12.918899} & Passive & PCA, LDA, BIC, LTP, HOG, TPLBP, FPLBP \& LBP used. & Intra-spectral (MWIR-MWIR) and inter-spectral (MWIR-Visible) matching performed.  \\ \hline
11 & Choi \textit{et al.}, 2012 \cite{doi:10.1117/12.920330} & Passive & LBP, HOG, Gabor filter and partial least squares-discriminant analysis & Used partial least squares-discriminant analysis to reduce the modality gap between thermal and visible images when cross-matching the two.  \\ \hline
12 & Osia \textit{et al.}, 2014 \cite{Osia:2014:SIA:2894850.2894944} & Passive & SIFT, SURF, PCA, LDA, BIC \& LBP used. & Development of a face identification system, where probe images are captured in visible or passive infrared bands. Direct matching is applied and the resulting system can work, both during day and night.  \\ \hline
13 & Reale \textit{et al.}, 2014 \cite{7025065} & Passive & DoG, HOG, PCA, SVM \& NN used. & Performed inter-spectral matching (Thermal-Visible) by learning coupled dictionaries.  \\ \hline
14 & Gurton \textit{et al.}, 2014 \cite{Gurton:14} & Passive & - & Use of polametric information to extract surface features of face.  \\ \hline
15 & Sarfraz \textit{et al.}, 2015 \cite{Sarfraz2015DeepPM} & Passive & SIFT \& PCA used.  & Application of a deep neural network for inter-spectral matching between thermal and visible images.  \\ \hline
16 & Short \textit{et al.}, 2015 \cite{Short:15} & Passive & DoG, HOG, RBF \& SVM  used. & Inter-spectral matching of LWIR and visible images by extraction of polametric information.  \\ \hline
17 & Hu \textit{et al.}, 2015 \cite{Hu16} & Passive & DoG \& HoG used. & Proposed Partial Least Squares for cross-matching and reducing the gap between thermal spectral signatures and visible spectral signatures.  \\ \hline
18 & Sarfraz \textit{et al.}, 2016 \cite{Sarfraz:2017:DPM:3086064.3086090} & Passive & DoG, SIFT \& PCA used.  & Application of a deep neural network for inter-spectral matching between thermal and visible images.  \\ \hline
19 & Iranmanesh \textit{et al.}, 2018 \cite{Iranmanesh2018DeepCP} & Passive & DoG used. & Deep learning methods used to match thermal images to visible images.  \\ \hline

%\end{tabular}
\caption{In chronological order, most important works in the passive IR domain} \label{PassiveIR}
\end{longtable}
\end{landscape}

%%%%%%%%%%%%%%%%%%
\begin{landscape}
\begin{longtable}{ | p{1cm} | p{4.5cm} | p{3cm} | p{5cm} | p{8cm} | }
%\centering
%\begin{tabular}
\hline

\textbf{SN} & \textbf{Author, Year} & \textbf{Modality type} & \textbf{Method(s)} & \textbf{Brief Description} \\ \hline

\endhead
\multicolumn{3}{@{}l}{\ldots \textit{continued on next page}}

\endfoot
\endlastfoot

1 & Chang \textit{et al.}, 2008 \cite{4563054} & Active/Passive & Fusion based on PCA and DWT & Catered for uncontrolled illumination using both thermal and active bands. \\ \hline
2 & Bourlai \textit{et al.}, 2012 \cite{6284307} & Active/Passive & FR using PCA, PCA+LDA, BIC, LBP \& LTP & Multiple scenarios considered: inter-spectral matching, different standoff distances, controlled \& uncontrolled conditions. \\ \hline

%\end{tabular}
\caption{In chronological order, work presented in both, active \& passive domains} \label{activepassive}
\end{longtable}
\end{landscape}

%%%%%%%%

\begin{landscape}
\begin{longtable}{| p{1cm} | p{4.5cm} | p{3cm} | p{5cm} | p{8cm} | }
%\centering
%\begin{tabular}
\hline

\textbf{SN} & \textbf{Author, Year} & \textbf{Modality type} & \textbf{Method(s)} & \textbf{Brief Description} \\ \hline

\endhead
\multicolumn{3}{@{}l}{\ldots \textit{continued on next page}}

\endfoot
\endlastfoot

1 & Chen  \textit{et al.}, 2005 \cite{chen2005ir} & IR, visible & PCA used for recognition & Inter-spectral matching and intra-spectral matching using Principal Component Analysis.  \\ \hline
2 & Chang \textit{et al.}, 2006 \cite{changfusion} & Visible sub-bands & PCA used. & A method proposed which takes into account variables from the spectral imaging system (such as skin reflectance, transmittance, etc.) and uses the these variables as weights.  \\ \hline
3 & Chang \textit{et al.}, 2008 \cite{4712365} & Visible sub-bands & - & Proposed a method to predict the optimal bands for recognition, fuse the images in those bands to obtain better performance over broadband images.  \\ \hline
4 & Chang \textit{et al.}, 2009 \cite{chang2009improving} & Multiple sub-bands & - & Selects the optimal bands for recognition, fuses the images to obtain better performance over broadband images.  \\ \hline
5 & Di \textit{et al.}, 2010 \cite{di2010studies} & Visible sub-bands & PCA used. & Decision level fusion strategy proposed.  \\ \hline
6 & Bouchech et al. \textit{et al.}, 2015 \cite{Bouchech:2015:KSA:2821861.2821895} & Visible sub-bands & LBP and its variants,  LGBPHS, HGPP, POEM-WPCA \cite{vu2012enhanced} \& Chang's algorithm \cite{chang2009improving}  used. & Face identification in unconstrained environment. Fusion applied to the best selected subbands of visible light  \\ \hline
7 & Chen \textit{et al.}, 2016 \cite{chen2016hyperspectral} & Multiple sub-bands & log-polar Fourier features extracted \& CRC used. & Use of image denoising methods to reduce noise in each image produced by a particular band.  \\ \hline
8 & Sharma \textit{et al.}, 2016 \cite{sharma2016image} & Multiple sub-bands & PCA, DCT, LBP, Gabor wavelets, HOG, LDP \& SIFT used. & Performance of Scale-Invariant Feature Transform evaluated against other descriptors for face recognition in hyperspectral images.  \\ \hline
9 & Jing \textit{et al.}, 2016 \cite{jing2016multi} & Multiple sub-bands & MLSDL used. & Used multi-view dictionary learning in hyperspectral sub-bands.  \\ \hline

%\end{tabular}
\caption{In chronological order, most important works in different sub-bands} \label{subbands}
\end{longtable}
\end{landscape}
%\twocolumn 

%%%%%%

\section{Identification of Spoof Attacks via Spectral Imaging} \label{spoofsection}

\begin{figure}
\centering
\includegraphics[scale=0.65]{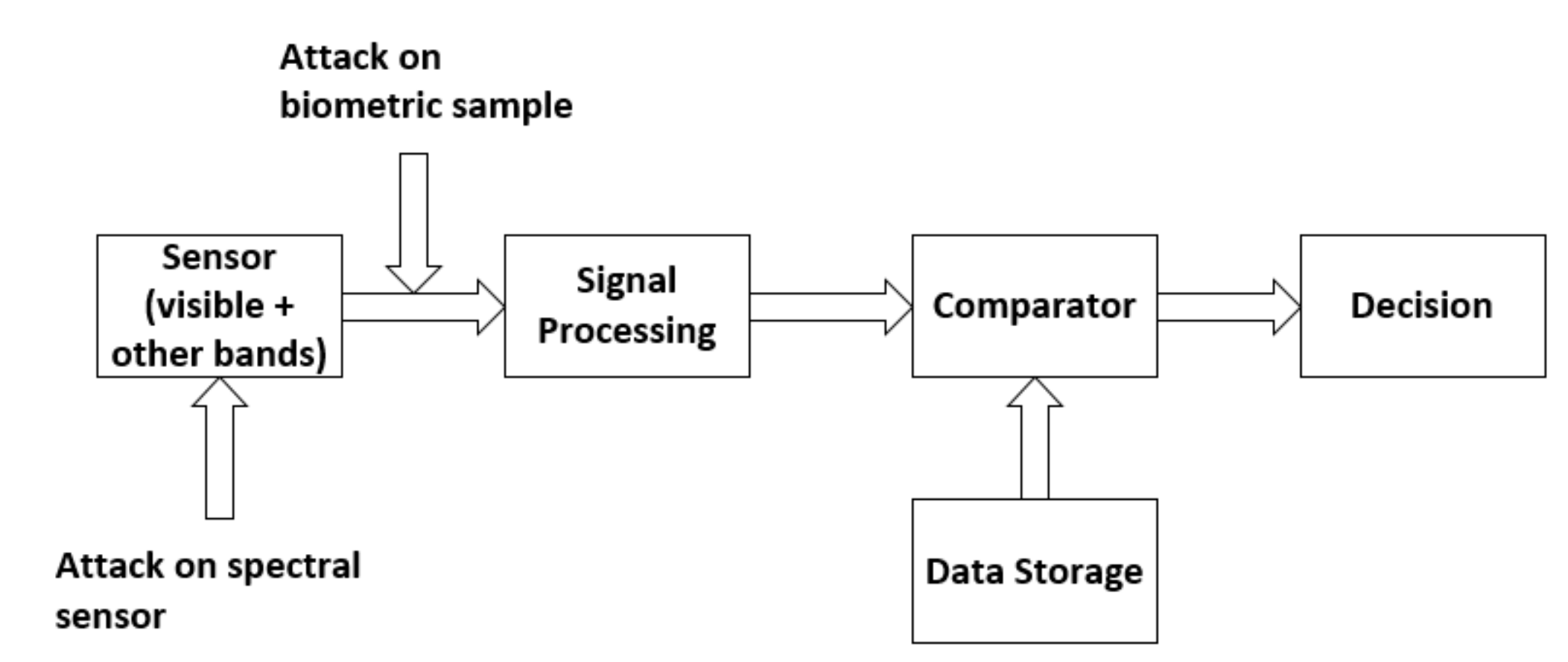}
\caption{A face recognition system with vulnerability at the sensor, inspired by Ramachandra \textit{et al.} \cite{Ramachandra:2017:PAD:3058791.3038924}} \label{huf}
\end{figure}
 
In the last section, our survey dealt with challenges for an operational biometric system using spectral sensors. This section deals with a general issue in all biometric systems when they are subjected to \textit{presentation attacks} defined below.

Current biometric systems for face recognition suffer from fraud, in the form of \textit{presentation attacks} or \textit{spoofing} \cite{Ramachandra:2017:PAD:3058791.3038924, 8244390, 7194916, 6909967}. In the context of biometrics, \textit{spoofing} can be defined as an attempt to gain access to the services of a system by use of biometric artifacts of a legitimate user. Also called a \textit{direct attack} or \textit{presentation attack}, these attacks affect the reliability of biometric systems, since they take place outside of the system. \textit{Indirect attacks} are those that happen inside the digital system such as an intruder gains access inside the system to alter the templates \cite{8244390}. A \textit{direct attack} does not require technical expertise and is therefore, easy to carry out, whereas an \textit{indirect attack} requires technical expertise which is why it is not as common as a \textit{direct attack}. 

Spectral imaging is the capture of images in multiple bands. Since these attacks are carried out at the sensor operating in the visible range, a sensor operating in another band can give more cues regarding the artifact or disguise used to carry out the attack. Figure \ref{huf} shows the block diagram of a face recognition system in which a spectral sensor is used and sensor vulnerability is shown. This sensor can capture images in visible spectrum and other spectral bands. We also see that the system is vulnerable to attacks, such as that at the sensor and an attack on the biometric sample. In the former attack, an \textit{artifact} (a copy of the biometric patterns of a legitimate user) is presented to the spectral camera. This is an example of a presentation attack. The second attack, as per Figure \ref{huf}, involves replacement of a legitimate sample with a fake.

It can be seen that it is easier to attack a sensor by producing an artifact of the face. Both, the cost and complexity of producing an artifact are low. The attack is possible in both of these cases \cite{Ramachandra:2017:PAD:3058791.3038924}:

\begin{itemize}

\item Authentication: A legitimate user's \textit{artifact} is presented at the spectral sensor to gain access to the services offered by the system to the legitimate user.

\item Identification: Modified biometric patterns of the attacker are presented to conceal the identity.   

\end{itemize}

According to Ramachandra \textit{et al.} \cite{Ramachandra:2017:PAD:3058791.3038924}, a Presentation Attack Instrument can be defined as a biometric object that can be used to carry out a  spoof attack. It can either be artificial (full such as face video, 3D mask, 2D print or partial such as a face with occlusion) or could involve human characteristics such as surgery or change in facial expressions. Artificial face artifacts of a legitimate user enrolled in a biometric system can be produced. These can be in the form of 3D masks, disguises such as makeup, photo prints with an InkJet printer or laser jet printer and electronic display (face photo or video). In this section, we review the work that has been done to counter spoof attacks using systems based on spectral imaging. We note that the work is not sufficient enough to establish or justify the widespread deployment of spectral systems to counter spoof attacks. We also note the absence of enough samples for the test purpose and the unavailability of a large, publicly available dataset to address spoof attacks on face. At best, these systems can be deployed alongside face recognition systems operating in the visible spectrum to enhance their capabilities. 

\subsection{Experiments}

In this section, we examine the experiments that have been conducted to identify spoof attacks in presence of different artifact types. We take note that there is not enough work in this area to identify spoof attacks using spectral imaging in presence of many varieties of artifacts. The artifacts used so far are limited to 3-D face mask and printed images. There is also an absence of common datasets for experimental evaluation. 

\subsubsection{Artifact type: 3-D face mask}

Steiner \textit{et al.} \cite{steiner2016reliable} addressed the problem of face disguises using 3D face masks via SWIR imaging. They incorporated a skin detection method as proposed by Steiner \textit{et al.} \cite{Steiner2016} with current face recognition systems using the visible spectrum. This is because to replace a visible spectrum based face recognition system with an imaging system in another spectral band requires the use of newly acquired databases. A spectral imaging system could be used in conjunction with a visible spectrum based face recognition system to enhance its capabilities. In their work, Steiner \textit{et al.} made the following contributions:

\begin{enumerate}

\item Successful detection of spoofing attacks when face masks or disguises were used with the help of an SWIR camera. This was done in a cooperative scenario where cross-matching was applied because the gallery images were captured in the visible spectrum.  

\item At the preprocessing step, skin classification was applied to the input image. At pixel level, this classification resulted in 'skin regions' and 'non-skin regions'. 

\item An anti-spoofing method that checks for the authentication of the face by selecting certain areas of interest in the face image.  

\end{enumerate}

\begin{figure}
\centering
\includegraphics[scale=0.80]{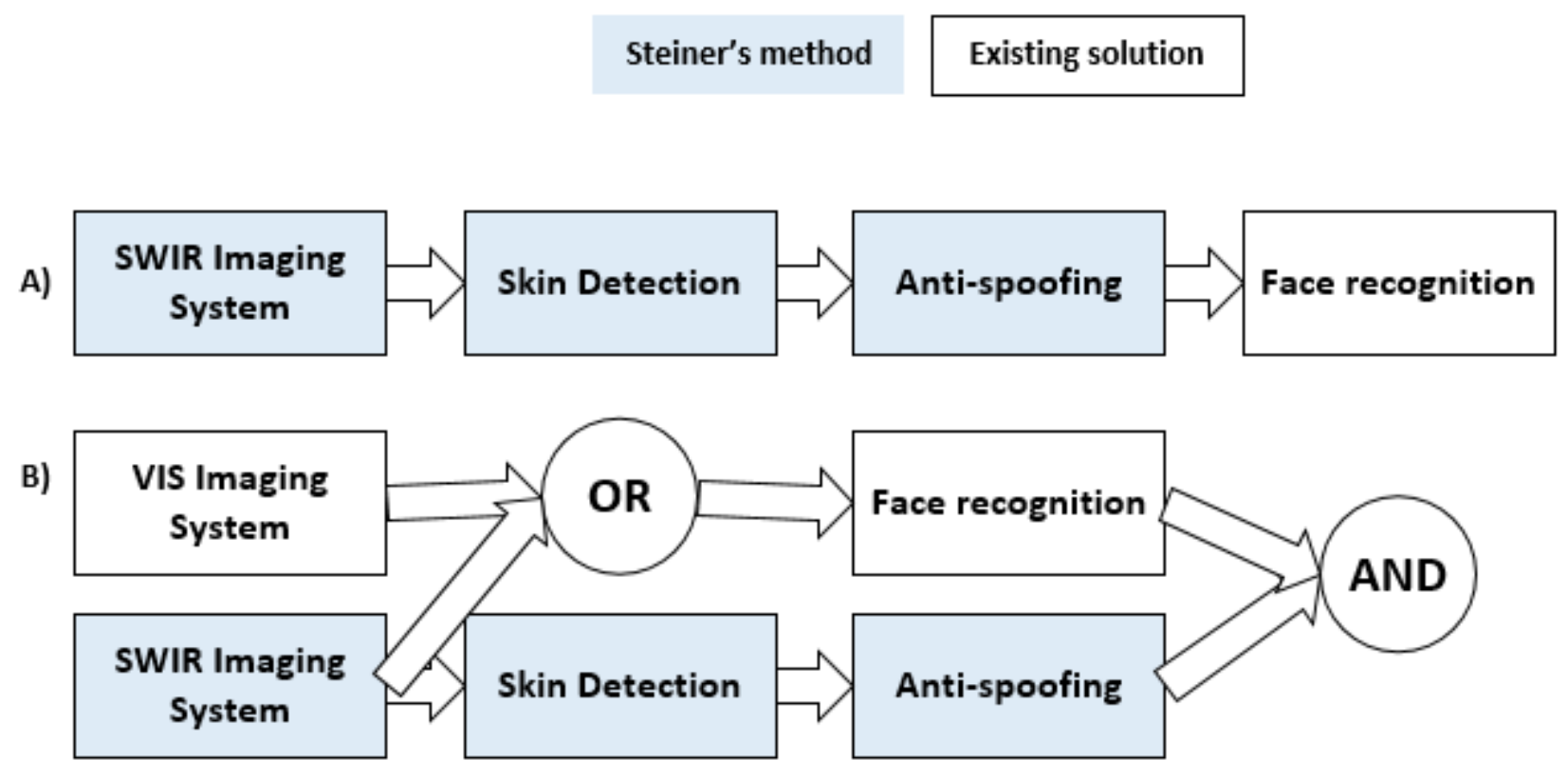}
\caption{Anti-spoofing method as proposed by Steiner \textit{et al.} \cite{steiner2016reliable}: (A) skin classification (B) ROI matching}\label{puf}
\end{figure}

The SWIR imaging system as proposed by Steiner \textit{et al.} \cite{Steiner2016}, uses 4 wavelengths (935 $\eta$m, 1060 $\eta$m, 1300 $\eta$m and 1550 $\eta$m). In their anti-spoofing method, two different skin detection methods (as seen in Figure \ref{puf}) have been used:

\begin{enumerate}

\item The face recognition module accepts the SWIR images on which skin classification is first applied. This classification separates skin regions and non-skin regions at pixel level and therefore, the method is termed as masking method. This is the preprocessing step before the SWIR image is provided as input to the face recognition module. 

\item An image in the visible spectrum is used as input to the face recognition module, in addition to the SWIR image. Skin classification is applied to the SWIR image and regions of interest (ROI) taken into account. This can be seen in Figure \ref{puf} where the authenticity of the SWIR image is verified by the anti-spoofing block. The result is then combined with that of the face recognition module. The system accepts the result if both blocks verify the face image. 

\end{enumerate}

Besides the SWIR sensor and an anti-spoofing module, a commercial software called Cognitec FaceVACS version 8.9 has been used for face recognition. A skin classifier based on Support Vector Machine (SVM) \cite{Steinwart:2008:SVM} has been used. It has been trained on authentic skin samples and different material samples such as makeup disguises and masks. BRSU Multispectral Skin/Face Database has been used which is publicly available \cite{brsudataset}. These cross-matching methods have been evaluated on the database which contains images of 137 subjects. 

When an attacker attempts to take the identity of another individual, it is a counterfeit attack. When such an attack occurs and the system verifies the person falsely, a false acceptance takes place. Steiner \textit{et al.} try a number of such attacks with full and partial masks. Both methods proposed by Steiner \textit{et al.}achieved a false acceptance rate of 0\%.

\subsubsection{Artifact type: InkJet Printer image \& Laser Jet Printer image}
% Raghavendra et al. collected an extended multispectral database of face images. It consisted of 50 subjects who were captured at a distance of 1.5-2 m from the multispectral camera, in seven sub-bands (425nm, 475nm, 525nm, 570nm, 625nm, 680nm and 930nm) over a period of 4 months in two sessions (5 images in each session). Both, bona fide and artifact images were captured using the same multispectral camera. Two sorts of presentation attacks were generated. The first one was generated using inkjet printer while the second one was generated using laser printer. In the case of laser print artifact, the multispectral device turned out to be vulnerable with the sub-bands 425nm, 475nm, 525nm, 625nm an 680nm. Low vulnerability was detected for the case of 930nm with SFAR= 1.25\% whereas vulnerability for the case of 680nm was high with SFAR= 100\%. For the case of inkjet print artifact, vulnerability was observed to be high on all sub-bands. The spectral sub-band of 680nm had the highest vulnerability with SFAR = 100\%.

Raghavendra \textit{et al.} \cite{raghavendra2017face} proposed a scheme to use spectral imaging to detect low-cost presentation attacks. An initial experimental evaluation had been carried out by Raghavendra \textit{et al.} \cite{raghavendra2017vulnerability}. This evaluation resulted in a low error rate on 930 $\eta$m band while remaining 6 sub-bands produced high error rates. 

The spectral signature of the captured image has certain, unique reflectance properties. In their work, when a sample (an artifact or a bona-fide) is presented to the spectral sensor, the spectral images are captured in various wavelengths. Spectral signature is extracted from each one of these images and a classifier is used to make a decision on the that signature. The reflectance properties of human skin and the artifact are different therefore, the proposed scheme can detect a presentation attack successfully. There has been work in literature earlier, to develop algorithms which successfully detect presentation attacks \cite{Ramachandra:2017:PAD:3058791.3038924, 7194916}. In the visible spectrum, these techniques, very accurately detect these attacks when presented with known artifacts. However, in a real world situation, the type of presentation attack is unknown. The use of a spectral imaging system can cater to such a situation where the attack is unknown. Raghavendra \textit{et al.} \cite{raghavendra2017face} used spectral signature acquired from images captured by the spectral sensor to successfully detect a presentation attack. They compared the proposed spectral approach to conventional methods such as image fusion. They also trained their algorithm to successfully detect unknown presentation attacks. 

\begin{figure*}
\centering
\includegraphics[scale=0.65]{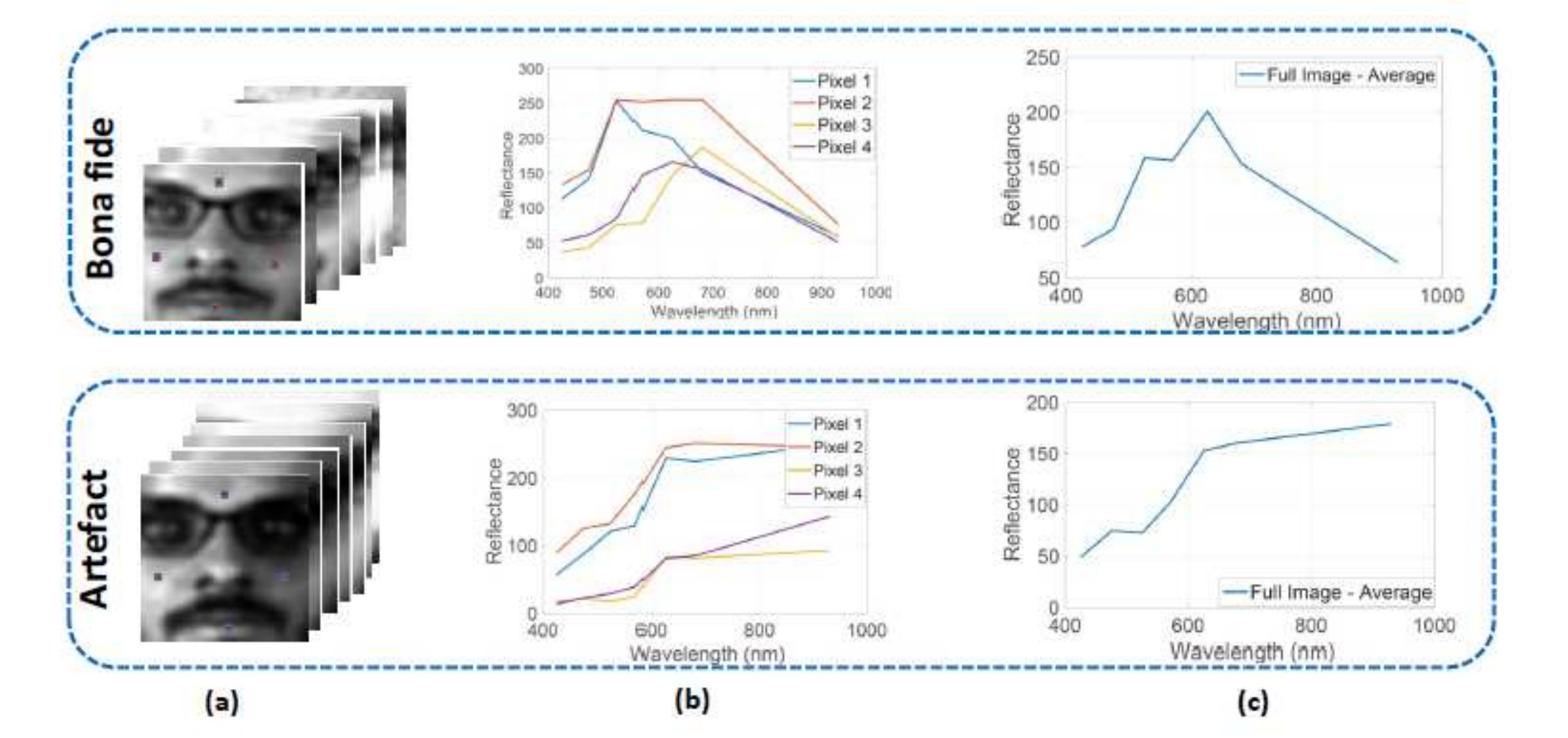}
\caption{Figure reproduced from \cite{raghavendra2017face} (a) Image acquired by a SpectraCamTM (b) spectral signature of 4 pixels (c) average spectral signature } \label{ggf}
\end{figure*}

A Multispectral camera - SpectraCamTM is used to acquire images in seven spectral bands (425 $\eta$m, 475 $\eta$m, 525 $\eta$m, 570 $\eta$m, 625 $\eta$m, 680 $\eta$m and 930 $\eta$m). In Figure \ref{ggf}(a), there are multispectral face images captured by SpectraCamTM, \ref{ggf}(b) spectral signatures extracted from four different pixels of the face which help in classifying as bona-fide or artifact and \ref{ggf}(c) gives the average spectral signature. What has been observed from Figure \ref{ggf}(b) and \ref{ggf}(c), is the discrimination in the spectral signatures of bona-fide and artifact. To classify the extracted features as bona-fide or artifact, Support Vector Machine (SVM) \cite{Steinwart:2008:SVM} is applied. Binarized Statistical Image Features (BSIF) \cite{bsif:kannala} and SVM have been used as methods to detect presentation attacks on the system.

The database used is called EMSPAD which stands for Extended Multispectral Presentation Attack Face Database. This database consists of images acquired for 50 subjects in two settings. 5 images have been acquired in each setting in 7 spectral bands. The bona-fide dataset consists of 3500 images. Two types of artifacts have been produced: (a) InkJet printer and (b) LaserJet printer. Canon EOS 550D DSLR has been used in 2 settings to capture the images. 5 images per setting have been acquired which were then used to produce the artifacts via printing. The artifact dataset consists of 3500 inkjet images and laserjet images each. Two partitions of the dataset were made: Partition-I consisted of 10 subjects while Partition-II consisted of 40 subjects. The first partition has been used to figure out the threshold value for classification of both, bona fide and artifact. The second partition has been used for testing and results. The second partition has been further divided into two sets each, consisting of 20 subjects. The first set is used for training and the second is used for testing. 

Two experiments have been conducted. In the first experiment, performance of the proposed technique has been evaluated for known attacks. Same artifacts are used for testing and training, each with laser print artifact and inkject print artifact. In the second experiment, performance of the proposed technique is cross-evaluated for unknown attacks. Therefore, training is first done with laser print artifact and testing with inkject print artifact. While in the second case, vice versa protocol is applied. The proposed scheme has given the best performance in all four experimental protocols compared to the conventional image fusion and score fusion methods. The dataset EMSPAD is publicly available as detailed by Raghavendra \textit{et al.} \cite{raghavendra2017vulnerability}. Inclusion of face spoofing datasets is out of the scope of this paper.
 
\section{Technological Advancements in Spectral Face Image Classification} \label{neuralsection}

Our previous section dealt with spoofing in FR systems. Very recently, deep learning has been used to deal with spoofing. Liu \textit{et al.} \cite{liu2018learning} combined Convolutional Neural Network and Recurrent Neural Network architectures to learn the depth in face images. They also introduced an anti-spoofing face database to perform experiments with their proposed method. However, this work dealt with images in the visible spectrum. In this section, we investigate how deep learning methods have been used incorporating spectral images.

According to LeCun \textit{et al.} \cite{lecun2015deep}, \textit{deep learning allows computational models that are composed of multiple processing layers to learn representations of data with multiple levels of abstraction}. The models learn from patterns or features from the data and propose to train the classifier by use of \textit{backpropagation algorithm}, in which a machine alters its parameters such that a computation for representation performed in one layer comes from that in the previous one. These methods have helped in advancements in various domains such as object recognition and detection, medical sciences, computer vision and natural language processing. 

Prior to the breakthroughs brought by deep learning-based algorithms, computer vision systems performed two tasks. In the first one, features were extracted from data by use of hand-crafted operators such that it would be easy to classify it. The second task involved using these features to train a classifier. This trained classifier would then be used to recognize an incoming probe to a high degree of accuracy. 

Feature extraction methods in the literature of spectral imaging for face recognition constitute the use of Gabor wavelets where with the help of magnitude and phase of the wave vector, scale and orientation is determined. Local operators are applied to the magnitude and phase, for instance, WLD and LBP are applied to the magnitude while GLBP is applied to the phase. This can be seen in \cite{robustnicolo, 6272346}. Cao \textit{et al.} proposed CMLD \cite{doi:10.1117/12.2176462} which combined Gabor filters, WLD, LBP and GLBP. Liang \textit{et al.} made use of 3D LDP as seen in \cite{liang20153d} while Uzair \textit{et al.} made use of 3D DCT \cite{uzair2013hyperspectral}. 

Deep learning based algorithms made many breakthroughs. These algorithms work by learning from data. Both feature extraction and classification are learned by the algorithms from the data. A Convolutional Neural Network (CNN) is a widely used deep learning architecture. It is capable of learning features from the images. It has been used for object detection in videos and images, for classification in large datasets and in remote sensing where hyperspectral imaging is used to capture data \cite{8068943}. 

\begin{figure*}
\centering
\includegraphics[scale=0.85]{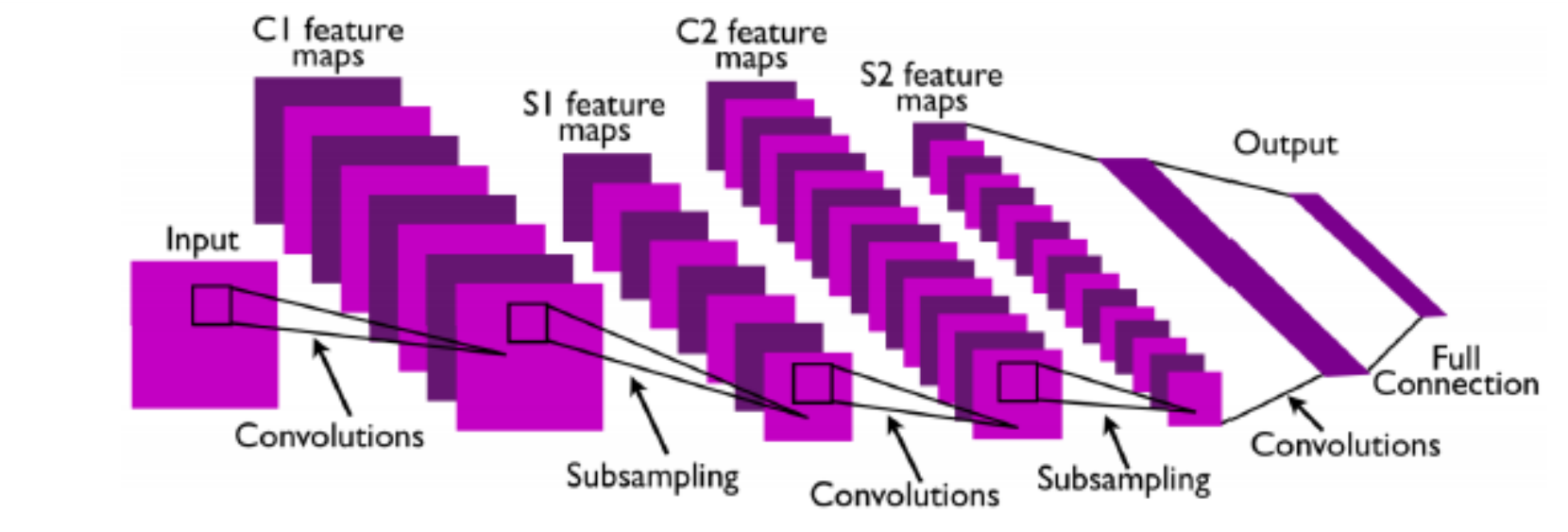}
\caption{A CNN architecture. Figure reproduced from \cite{lecun2010convolutional}} \label{CNN}
\end{figure*}

The architecture of a CNN, as seen in Figure \ref{CNN}, was first proposed by LeCun \cite{lecun2010convolutional}. It is a multi-stage or multi-layer architecture. This essentially means there are multiple stages in a Convolutional Neural Network for feature extraction. Every stage in the network has an input and output which is composed of arrays known as feature maps. At a certain stage, every output feature map consists of patterns or features extracted on locations of the input feature map. Every stage is made up of three layers after which classification takes place. These layers are \cite{corr1803, zeilerconv}:

\begin{enumerate}

\item Convolution layer: This layer makes use of filters, which are convolved with the image producing activation or feature maps. 
\item Rectification layer: This layer makes use of non-linear functions to get positive activation maps.
\item Feature Pooling layer: This layer is inserted to reduce the size of the image representation to make the computation efficient. The number of parameters is also reduced which in turn controls over-fitting. 

\end{enumerate}

CNNs have shown to give a superior performance than conventional methods. However, the use of CNNs for spectral image classification undergoes through overfitting, due to scarcity of enough sample points. This setback has also been highlighted in \cite{lezama2017not}, where Lezama \textit{et al.} acknowledge how breakthroughs have been made with visible images by application of deep learning methods but the same has yet to be seen in spectral images. They argue that this is due to the absence of enough training samples for spectral images. 

In the subsequent sections, we see how deep learning has been used in the domain of spectral imaging, the frontiers and the setbacks.  

\subsection{Face recognition using S-CNN architecture}

Sharma \textit{et al.} \cite{sharma2016hyperspectral} worked on a deep learning-based method for spectral imaging by training their own Convolutional Neural Network (CNN) architecture to learn discriminative features in InfraRed for better classification accuracy. Sharma \textit{et al.} used CNNs for classification of hyperspectral images and then used the extracted features for discriminative band selection. According to them, the content of a hyperspectral image is rich in information and can be harnessed for much better classification accuracy. However, the bulky hyperspectral cube is a setback because it makes computation exhaustive. The dimensionality of hyperspectral data is high. 

Sharma \textit{et al.} proposed a framework for classification of hyperspectral face images and band selection at image-level (instead of pixel-level), to extract specifically those wavelengths that were discriminative and uncorrelated. These selected bands would be rich in information and help in accurate classification. This is a fundamental issue for real-time biometric systems. If uncorrelated wavelengths are chosen, they give good face recognition rates apart from reducing redundancy, simplifying exhaustive computation and reducing the time required to acquire the cube. The proposed CNN model, called S-CNN, consists of 3 convolution layers, 2 fully-connected layers and a \textit{C}-way softmax. Every convolution layer has a batch-normalization layer, a rectification linear unit layer and a max pooling layer.

Sharma \textit{et al.} extracted hand-crafted features for performance comparison with their proposed CNN model. For this purpose, they extracted HOG, LBP and DSIFT-FVs features. AdaBoostSVM algorithm is applied to select best 3 bands from the hyperspectral cube for each feature. Features are fed to a multi-class linear SVM and then merged using AdaBoost. Two databases are used: CMU-HSFD \& HK PolyU-HSFD. The results are compared with those obtained through the use of CNN model proposed by Sharma \textit{et al.}. Table \ref{sharmaft} shows accuracy rates when all bands are chosen and no band selection is applied. We see that the proposed method outperforms other conventional methods. 

\begin{table*}[!h]
\centering
\begin{tabular}{ | l | l | l | }
\hline

\textbf{Method} & \textbf{PolyU-HSFD} & \textbf{CMU-HSFD}  \\ \hline
\textbf{3D-DCT} & 84 & 88.6  \\ \hline
\textbf{3D Gabor Wavelets} & 90.1 & 91.6  \\ \hline
\textbf{3D LDP} & 95.3 & 94.8  \\ \hline
\textbf{2D PCA} & 71.1 & 72.1  \\ \hline
\textbf{LBP} & 85.6 & 86.1  \\ \hline
\textbf{HOG} & 92.3 & 91.5  \\ \hline
\textbf{DSIFT-FVs} & 96.1 & 96.9  \\ \hline
\textbf{S-CNN} & 97.2 & 98.8  \\ \hline
\textbf{S-CNN+SVM} & 99.3 & 99.2  \\ \hline

\end{tabular}
\caption{This table shows percentage accuracy rates of various conventional methods compared to those of S-CNN, when all bands are used} \label{sharmaft}
\end{table*}

Best 3 bands are chosen by band selection to give the most uncorrelated bands, as seen in Table \ref{best3bands}. Recognition accuracy results are affected when best 3 bands are selected which are different for every operator. However, very high accuracy results are reported for the proposed S-CNN+AdaBoostSVM compared to other methods, as seen in Table \ref{sharmast}.

\begin{table*}[!h]
\centering
\begin{tabular}{ | l | l | l |  }
\hline

\textbf{Method} & \textbf{PolyU-HSFD} & \textbf{CMU-HSFD}  \\ \hline
\textbf{LBP} & [560, 650, 510] & [720, 730, 650]  \\ \hline
\textbf{HOG} & [460, 510, 570] & [1030, 1010, 1050]  \\ \hline
\textbf{DSIFT-FVs} & [460, 640, 520] & [630, 680, 690]  \\ \hline
\textbf{S-CNN+AdaBoostSVM} & [590, 570, 520] & [900, 970, 730]  \\ \hline

\end{tabular}
\caption{This table shows the best 3 bands obtained by band selection method} \label{best3bands}
\end{table*}

\begin{table*}[!h]
\centering
\begin{tabular}{ | l | l | l |  }
\hline

\textbf{Method} & \textbf{PolyU-HSFD} & \textbf{CMU-HSFD}  \\ \hline
\textbf{LBP} & 79.4 & 99.6  \\ \hline
\textbf{HOG} & 80 & 91.5  \\ \hline
\textbf{DSIFT-FVs} & 88.3 & 98.4  \\ \hline
\textbf{S-CNN+AdaBoostSVM} & 99.4 & 99.2  \\ \hline

\end{tabular}
\caption{This table shows percentage accuracy rates of various conventional methods compared to those of S-CNN, when best 3 bands are used} \label{sharmast}
\end{table*}

\subsection{NIRFaceNet}

Peng \textit{et al.} proposed NIRFaceNet \cite{peng2016nirfacenet} for face recognition using NIR image sets. NIRFaceNet is targeted towards non-cooperative user scenarios and is suited for surveillance and security applications. In a practical non-cooperative case, there is noise in the captured images due to the way in which these are captured. To cater for this, Peng \textit{et al.} used the CASIA NIR Database and added noise manually to the images of the database. They created 9 set of images with different types of noise or occlusion added to each. These are called test sets (test sets are labeled from 1 to 9). NIRFaceNet successfully achieved a very high identification rate with less training and processing times. In their model, Peng \textit{et al.} modified the GoogLeNet (a deep neural network with 22 parameter layers but with pooling, 27 layers \cite{szegedy2015going}) to make it suitable to work with smaller images' set compared to the ImageNet database which is very large. The architecture proposed by Peng \textit{et al.} consists of only 8 layers and has 2 modules for feature extraction. 

We report results obtained from applying NIRFaceNet, GoogLeNet, LBP+PCA, LBP Histogram \cite{KHAN20131159}, ZMUDWT and ZMHK to different test sets. The first set, test set 1, consists of 3 images of normal faces for all subjects. This test set excludes training sets and subjects' images that are less than 3. Test set 2 consists of all images of the subjects excluding those with glasses. This also excludes training set. In test set 3, motion blur is added to images in test set 2. In test set 4, Gaussian blur is added with standard deviation of 0.5, to test 2 images. In test set 5, Gaussian blur is added with standard deviation of 2, to test set 2 images. In test set 6, salt-pepper noise is added with density of 0.01 to images used in test set 2. In test set 7, salt-pepper noise with density of 0.1 is added to images used in test set 2. In test set 8, Gaussian noise with mean 0 and variance 0.001 is added to images used in test set 2. In test set 9, Gaussian noise with mean 0 and variance 0.01 is added to images used in test set 2. For all these sets, accuracy rates have been shown in Table \ref{facenetrates}. We see that NIRFaceNet outperforms all other methods in each test set.  

\begin{table*}[!h]
\centering
\begin{tabular}{ | l | l | l | l | l | l | l | l | l | }
\hline
	\textbf{Test Set ID} & \textbf{M1} & \textbf{M2} & \textbf{M3} & \textbf{M4} & \textbf{M5} & \textbf{M6} & \textbf{M7} & \textbf{M8} \\ \hline 
\textbf{Test Set 1} & 89.76 & 87.34 & 95.64 & 100 & 99.02 & 98.8 & 98.74 & 100 \\ \hline 
\textbf{Test Set 2} & 80.94 & 87.34 & 90.18 & 96.5 & 95.64 & 95.15 & 94.73 & 98.28 \\ \hline 
\textbf{Test Set 3} & 30.92 & 54.14 & 88.35 & 95.14 & 94.33 & 93.79 & 93.04 & 98.12 \\ \hline 
\textbf{Test Set 4} & 76.85 & 82.99 & 90.03 & 96.5 & 94.98 & 95.15 & 94.2 & 98.48 \\ \hline 
\textbf{Test Set 5} & 30.27 & 46.4 & 88.97 & 95.25 & 93.25 & 93.03 & 92.04 & 98.24 \\ \hline 
\textbf{Test Set 6} & 78.02 & 81.38 & 89.89 & 95.87 & 95.2 & 95.05 & 94.88 & 98.32 \\ \hline 
\textbf{Test Set 7} & 20.45 & 12.6 & 82.48 & 90.51 & 94.79 & 94.04 & 93.73 & 96.02 \\ \hline 
\textbf{Test Set 8} & 0.99 & 1.61 & 89.63 & 96.2 & 95.79 & 96.03 & 95.25 & 98.36 \\ \hline 
\textbf{Test Set 9} & 0.66 & 1.35 & 87.77 & 94.56 & 92.41 & 92.2 & 91.73 & 97.48 \\ \hline 

\end{tabular}
\caption{This table shows percentage accuracy rates obtained using NIRFaceNet and other methods. \textbf{M1}: LBP+PCA, \textbf{M2}: LBP Histogram, \textbf{M3}: ZMUDWT, \textbf{M4}: ZMHK, \textbf{M5}: GoogLeNet softmax0, \textbf{M6}: GoogLeNet softmax1, \textbf{M7}: GoogLeNet softmax2 and \textbf{M8}: NIRFaceNet} \label{facenetrates}
\end{table*}

\subsection{Transfer learning}

The NIRFaceNet was specifically designed to work for CASIA NIR Database. Therefore, it could be redesigned to include other publicly available NIR Databases. Building a large database is crucial for training purpose \cite{Khan2019}. A database consisting of images in other modalities needs to be produced to understand the merits of deep neural networks for face recognition in different modalities. However, to cater for the absence of a large database in other modalities, Lezama \textit{et al.} \cite{lezama2017not} proposed a transfer learning approach combined with joint embedding. Most deep learning models are trained for visible images. Lezama \textit{et al.} proposed to preprocess NIR images using CNN in such a way that converted the NIR images into visible images. Then this hallucinated visible image is presented as input to a DNN model. The CNN model used for producing the hallucinated images is trained in the YCbCr space. Another approach used was to apply low-rank embedding at the output of the model where it produces deep features for both, visible and NIR images in a common space. These two approaches were then combined as well to compare results. 

Lezama \textit{et al.} tried their approach using pre-trained DNN models for visible images. These are:

\begin{itemize}

\item VGG-S model: This model uses the VGG-S architecture.
\item VGG-face model: This is a publicly available DNN face model. 
\item COTS: This is a Commercial-Off-The-Shelf face model.  

\end{itemize} 

CASIA NIR-VIS 2.0 Face Database was used for experiments. We report rank-1 identification rates for cross-spectral hallucination and low-rank embedding using the pre-trained DNN models in the Table \ref{lezamadnn}. Combining cross-spectral hallucination and low-rank embedding outperforms all other cases, where either only hallucination is applied to the input of the DNN model or only low-rank embedding is applied to the output of the DNN model. 

\begin{table*}[!h]
\centering
\begin{tabular}{ | l | l | l | l |  }
\hline
	\textbf{Case ID} & \textbf{VGG-S} & \textbf{VGG-face} & \textbf{COTS} \\ \hline
\textbf{Case 1} & 75.04 & 72.54 & 83.84 \\ \hline
\textbf{Case 2} & 80.65 & 83.1 & 93.02 \\ \hline
\textbf{Case 3} & 89.88 & 82.26 & 91.83 \\ \hline
\textbf{Case 4} & 95.72 & 91.01 & 96.41 \\ \hline

\end{tabular}
\caption{This table shows rank-1 identifications rates obtained from pre-trained DNN models. \textbf{Case 1}: DNN model used as it is, \textbf{Case 2}: Hallucination is applied, \textbf{Case 3}: Low-rank embedding is applied, \textbf{Case 4}: Hallucination and low-rank embedding are combined} \label{lezamadnn}
\end{table*}

\section{Conclusion \& Future Work} \label{futuresection}

In this section, we review the hurdles that need to be overcome, in order to make use of biometric systems based on spectral imaging for face recognition. Some of these are:

\begin{enumerate} 

\item \textbf{Sensor cost:} Development of low-cost spectral imaging systems for face recognition has been a hurdle. One such system has been developed by Vetrekar \textit{et al.}, in literature. 9 narrow-band filters were used to capture images at wavelengths 530 $\eta$m, 590 $\eta$m, 650 $\eta$m, 710 $\eta$m, 770 $\eta$m, 830 $\eta$m, 890 $\eta$m, 950 $\eta$m and 1000 $\eta$m. The system makes use of a CMOS sensor (model: BCi5-U-M-40-LP) which is cheap, consumes less power and is smaller in size \cite{7738245}. However, there is a need to implement the same for practical use. 

\item \textbf{Database availability:} The unavailability of large publicly available hyperspectral/multispectral face databases for the purpose of testing and experimentation is a setback for the research community. We have observed that the publicly available databases are very few in number. There is no large database for benchmark evaluations. The existing databases have not captured the same test subjects in all possible bands in which work has been done so far, for experimental evaluation. Though results have been encouraging however, there has to be a common evaluation database for a large number of test subjects. It's not possible to cross evaluate the currently available databases. There is a need to use the same wavelengths to capture a large dataset to be used for benchmarking of all methods. 

\item \textbf{Limitation of Passive IR:} The use of thermal imaging, though makes spectral signatures unique is affected by the ambient temperature, exercise, stress, sickness, etc. as highlighted in \cite{ghiass2014infrared}. Since the thermogram produced by thermal imaging is highly sensitive to external factors, it is very challenging to use this technology in conjunction with current systems.

\item \textbf{Limitation of Active IR:} The use of NIR and SWIR though seems plausible for surveillance activities since they are resistant to the presence of occlusions or pose changes and illumination conditions, however, they suffer from sensitivity to sunlight. The high recognition rates reported in most of the earlier works take into account, cooperative and controlled conditions while accuracy degrades in case of non-cooperative and uncontrolled conditions where variable stand-off distances and light changes are taken into account. 

\item \textbf{Artifacts:} Though there has been work in literature on detection of spoof attack using spectral imaging, the different artifacts to carry out such an attack lack variety. In a practical situation, the attack is unknown. Therefore, more experiments need to be conducted to set ground for use of spectral imaging to counter direct attacks.

\item \textbf{Exploitation of deep learning-based methods:} This has been limited so far due to small number of test subjects and their images in the existing databases. Deep learning-based methods require large number of parameters for training. Since the number of such samples is low in the existing databases, this leads to over-fitting.  

\end{enumerate}

We realise that with these issues related to spectral imaging still at large, the wide-spread deployment of spectral sensors for face recognition will remain hampered for sometime. Therefore, researchers need to initially acquire a large publicly available dataset for spectral imaging in all modalities (Active \& Passive) and in wavelengths that have performed well, as step one, for extensive experiments and cross-evaluation of methods and algorithms. 

However, the technology shows promise when used in conjunction with visible imaging systems. Spectral imaging systems have shown to provide higher recognition rates in the presence of occlusion, to counter spoof attacks and facial expression variations \cite{farokhi2014near, farokhi2015near}. Visible imaging systems have not been able to cater to these issues so far, but can make use of spectral imaging systems. To cater for the absence of a single, large database, recently transfer learning has been used by Lezama \textit{et al.} \cite{lezama2017not} which has shown promise in face of scarce number of available image samples to train deep learning models. This is a way forward for the deep learning community to use existing available spectral datasets for development of deep learning models that can give a lead towards task-specific and application-specific recognition cases.

%\bibliography{References_SpecImag}

%%%%%%%%%%%%%%%%%%%%%%%%%%%%%%%%%%%%%%%%%%%%%%%%%%%%%%%%%%%%%%%%%%%%%%%%%%

%%%%%%%%%%%%%%%%%%%%%%%%%%%%%%%%%%%%%%%%%%%%%%%%%%%%%%%%%%%%%%%%%%%%%%%%%%

\end{document}